\documentclass[journal]{IEEEtran}
\usepackage{cite}
\usepackage{array}
\usepackage{mdwmath}
\usepackage{mdwtab}
\usepackage{eqparbox}
\usepackage{subfigure,graphicx}
\usepackage{enumerate}
\usepackage{epstopdf}
\usepackage{multirow}
\usepackage{algorithm, algpseudocode, algorithmicx}
\usepackage{color}
\usepackage{epstopdf}
\usepackage[switch]{lineno}
\usepackage{booktabs}
\usepackage{diagbox}
\usepackage{float}
\usepackage{amsmath}
\usepackage{geometry}
\usepackage{url}

\newcommand{\tabincell}[2]{\begin{tabular}{@{}#1@{}}#2\end{tabular}}
\makeatletter
\def\hlinew#1{%
  \noalign{\ifnum0=`}\fi\hrule \@height #1 \futurelet
   \reserved@a\@xhline}
\makeatother

\begin{document}

\title{{Deep Learning-Based Gait Recognition Using Smartphones in the Wild}}

\author{Qin~Zou,
        Yanling Wang,
        Qian~Wang,
        Yi Zhao,
        Qingquan Li\\
        $\ $ \\
        {\color{blue}\url{https://github.com/qinnzou/Gait-Recognition-Using-Smartphones}}

\thanks{Q.~Zou and Y.~Zhao
are with the School of Computer Science, Wuhan University, Wuhan 430072,
P.R.~China (E-mails: qzou@whu.edu.cn, yizhaowhu@whu.edu.cn).}
\thanks{Y.~Wang and Q.~Wang
are with the School of Cyber Science and Engineering, Wuhan University, Wuhan 430079,
P.R.~China (E-mails: yanling@whu.edu.cn, qianwang@whu.edu.cn).}
\thanks {Q.~Li is with Shenzhen Key Laboratory of Spatial Smart Sensing and Service,
Shenzhen University, Guangdong 518060, P.R.~China (E-mail: liqq@szu.edu.cn).}

}

\markboth{IEEE Transactions on Information Forensics and Security, 2020}
{Shell \MakeLowercase{\textit{et al.}}: }

\maketitle

\begin{abstract}
Compared to other biometrics, gait is difficult to conceal and has the advantage of being unobtrusive. Inertial sensors, such as accelerometers and gyroscopes, are often used to capture gait dynamics. These inertial sensors are commonly integrated into smartphones and are widely used by the average person, which makes gait data convenient and inexpensive to collect. In this paper, we study gait recognition using smartphones in the wild. In contrast to traditional methods, which often require a person to walk along a specified road and/or at a normal walking speed, the proposed method collects inertial gait data under unconstrained conditions without knowing when, where, and how the user walks. To obtain good person identification and authentication performance, deep-learning techniques are presented to learn and model the gait biometrics based on walking data. Specifically, a hybrid deep neural network is proposed for robust gait feature representation, where features in the space and time domains are successively abstracted by a convolutional neural network and a recurrent neural network. In the experiments, two datasets collected by smartphones for a total of 118 subjects are used for evaluations. The experiments show that the proposed method achieves higher than 93.5\% and 93.7\% accuracies in person identification and authentication, respectively.
\end{abstract}


\IEEEpeerreviewmaketitle

\section{Introduction}

\IEEEPARstart{B}{iometrics} refers to the automatic identification of a person based on his or her physiological or behavioral characteristics. With the increasing demand for person identification and verification in the era of big data and artificial intelligence, the research and development of biometric systems have attracted broad attention from both academia and industry. Many biometrics, e.g., the fingerprint, iris, face, and voice etc., have been implemented commercially.  Some of these biometrics are obtrusive to users as they require the cooperation of users to collect the data. For example, users are asked to place a finger on a device to have their fingerprints captured or to look at a camera close enough to have their irises imaged. In such cases, a user may feel offended and easily realizes that his/her identity is being checked. Meanwhile, some biometrics are easily forged and attacked. For example, face recognition can be cheated by using an image or a video of the target face~\cite{tronci2011fusion,kim2013face}. As a result, less obtrusive and more robust biometrics are currently in great demand.

Unobtrusiveness is especially important for a biometric system that must work in a discrete manner, e.g., to recognize the identity of a person but not to let him/her know he/she is being identified. Among various biometrics, gait not only satisfies the requirement of being unobtrusive but also is more difficult to conceal~\cite{Nixon2006,zhang2010tsmc,ding2015tc,minaee2019biometric}.

Gait biometric involves identifying a person based on his/her walking characteristics. Generally, gait recognition can be performed on two types of data: a sequence of images (e.g., from a video), or an inertial gait time series generated by inertial sensors.
If gait images or inertial gait time series can be captured, we can perform gait recognition and hence person identification.

For vision-based methods, Osaka University provided the very early and effective work~\cite{makihara2006gait}, where the gait data is collected on thousands of subjects. With the development of advanced semantic segmentation algorithms~\cite{lin2017refinenet,he2017mask}, the segmentation of a walking person from an arbitrary background has become possible. Even in cluttered backgrounds, accurate silhouette can be extracted~\cite{zhang2019gait}, which leads to the successful gait recognition in complex environments~\cite{rida2018IET}. Besides, some studies bypass the use of silhouette images~\cite{liao2017pose,an2018improving,zhang2019comprehensive}, e.g., using pose estimation.

Inertia-based methods, inertial sensors, such as accelerometers and gyroscopes, are used to record the inertial data generated by the movement of a walking body. These inertial data capture the gait dynamics in a general manner and have been shown to be useful for extracting walking patterns~\cite{zou2017robust}. In the past decade, multiple methods for inertia-based gait recognition have been developed~\cite{gafurov2009gait,XuCMU12btas,Trung12icb,zhang2015tc,sprager2015efficient}. However, most of them require inertia sensors to be fastened to specific joints of the human body, which is inconvenient for the gait-data collection.

In real-world application scenarios, the vision-based gait recognition is often used when closed circuit televisions (CCTVs) are installed around the target subjects, which may be unaware of the observation. Inertia-based approaches, however, can be adopted even without the existence of CCTVs, e.g., electronic shackles are put on a prisoner for supervisory control.

As known, many advanced inertial sensors, including accelerometers and gyroscopes, are commonly integrated into smartphones nowadays~\cite{shen2020pattern,shen2018activeauth}. So it is very convenient and cheap to collect the interial gait data, and these benefits have inspired a number of solutions to adopt smartphones for gait recognitions~\cite{Kwapisz10icb,sun2014gait,fernandez2017optimizing}. Smartphone-based gait recognition has many demands, e.g., person identification and authentication. To achieve person identification with high accuracy, most existing methods require the person to walk along a specified road (e.g., a straight lounge) and/or at a normal speed. Obviously, such strict requirements heavily limit its wide application, which motivates us to design more robust gait-recognition algorithms with loose constraints.

In recent years, deep learning has achieved human-competitive, and sometimes better-than-human performance in solving many cognitive problems such as speech recognition~\cite{graves2013speech,ZhouWYLJCW19} and visual perception~\cite{krizhevsky2012imagenet,donahue2015long,Zou2018deepcrack,chen2019deep,shao2019saliency,zhang2019improved,shao2019cloud} etc. There are mainly two types of deep neural networks: the deep convolutional neural network (DCNN) and the deep recurrent neural network (DRNN)~\cite{zou2019tvt}. The former convolves input signals in the space domain and is suitable for handling array signals, e.g., images. The latter processes input signals in a recursive manner and is suitable for handling time series, e.g., voices. The data produced by accelerometers and gyroscopes can naturally be arranged into a two-dimensional time series. Hence, DCNNs can represent inertial data by convolutional feature maps, and DRNNs can exert their advantages by processing the data as time series. However, how to combine the two types of neural networks for effective gait representations remains unknown.

In this paper, we study gait recognition using smartphones in the wild and propose an effective and seamless combination of DCNN and DRNN for robust inertial gait feature representation. During the gait data collection, smartphones are assumed to be used under unconstrained conditions, and they do not record when, where, and how a user walks. Under this assumption, the inertial data collected by smartphones is partitioned into walking sessions and non-walking sessions by means of a fully convolutional neural network, where hierarchical convolutional features are fused to accurately extract walking sessions. Then, gait features are extracted from the walking data by our proposed hybrid deep learning technique. Specifically, the three-dimensional data (along the X, Y and Z axes) of the accelerometer and gyroscope are combined to form a six-dimensional time-series. Then, a CNN with one-dimensional kernels is designed to convolve the input time series into convolutional feature maps, which can maintain the time-series characteristics. Next, the time-series convolutional features are processed by an RNN for robust gait feature extraction. Based on the above operations, person identification and authentication models can finally be constructed in a supervised training manner. Two datasets are constructed to evaluate the proposed method in person identification and authentication.

The main contributions of this paper are twofold:

\begin{enumerate}[    $\vcenter{\hbox{\tiny$\bullet$}}$]

\item First, to address the problems that traditional methods cannot achieve high performance in inertia-based gait recognition and that DCNNs and DRNNs often function separately in different domains, a deep-learning architecture is designed to seamlessly integrate the two networks for gait feature representation. Specifically, a DCNN, with a sequence of carefully designed one-dimensional kernels, guarantees that the output convolution features maintain the property of time series; then, a long short-term memory (LSTM) network processes the resulting features for gait recognition.

\item Second, to evaluate the performance of smartphone-based gait recognition in unconstrained conditions, two main datasets are collected. One dataset is collected from 20 subjects, with each subject providing thousands of samples. The other dataset is collected from 118 subjects, with each subject providing hundreds of samples. Based on the two datasets, six sub-datasets are constructed and used for quantitative evaluation and performance comparison.
\end{enumerate}

\section{Related Work}\label{sec:relate}
\subsection{Gait Recognition Using Inertial Sensors}
Sensor-based gait recognition can be performed in three main ways: by sensors in the floor~\cite{addlesee1997orl}, by sensors in the shoes~\cite{huang2007gait}, and by sensors on the body~\cite{gafurov2009gait}. Among these methods and their variations, inertia-sensor based methods are the most attractive. This is because that the inertial sensors can be easily placed on the body to capture the details of the movement characteristics~\cite{plotz2011feature,yang2008using,alsheikh2016deep,ward2006activity,shrestha2015tap,mayagoitia2002accelerometer,zhao2018heading}, and the captured time-series gait data are effective for person identification and authentication~\cite{frank2010activity,sprager2015efficient}.

In early research of inertia-based gait recognition, Ailisto {\it et al.} proposed a signal-correlation method, where the recognition was performed in the means of template matching and cross-correlation computation~\cite{mantyjarvi2005identifying,ailisto2005identifying}. Following this work, Gafurov {\it et al.} made many significant improvements~\cite{gafurov2006biometric,gafurov2007spoof,gafurov2007gait,gafurov2009gait}. In~\cite{gafurov2007spoof}, they analyzed the minimal-effort impersonation attack and the closest person attack on gait biometrics. In~\cite{gafurov2007gait}, they collected 300 gait sequences from 50 subjects by placing an accelerometer sensor in the user's pocket, and achieved an equal error rate (EER) of 7.3\%. In~\cite{gafurov2009gait}, they tried foot-, pocket-, arm- and hip-based user authentication and found that a sideways motion of the foot provides the most discrimination, and a different segment of the gait cycle often leads to a different level of discrimination.

Beside the above work, many other gait-recognition methods have been developed since 2007. In~\cite{Liu07icbb}, Liu {\it et al.} employed the dynamic time warping (DTW) as a tool for gait-curve matching. This work was improved in~\cite{Liu07iciea}, where the wavelet denoising and gait-cycle segmentation algorithms were introduced for data preprocessing. In~\cite{trivino2010application}, Trivino {\it et al.} proposed an approach using a fuzzy finite state machine (FFSM) to model the perception of the signal evolution, which achieved superior results to that of Gafurov and Liu. In~\cite{zhang2015tc}, Zhang {\it et al.} proposed an effective algorithm that avoided cycle-detection failures and inter-cycle phase misalignment. In~\cite{Derawi10iih}, Derawi {\it et al.} provided a stable cycle detection mechanism and improved the gait-based authentication. In~\cite{sprager2015inertial}, an overview of the inertia-based gait recognition methods was given with extensive comparisons.

Another representative work was conducted by a research team from Osaka University~\cite{Trung11ijcb,Trung12icb,ngo2014orientation}. In their sensor-based approach, they have provided the largest inertia-based gait dataset in the world, including 744 subjects (389 males and 355 females) with ages ranging from 2 to 78 years~\cite{Ngo14pr}, which is a significant contribution to the community~\cite{zhong2014sensor}. In their study, the performance of gait recognition based on acceleration was found to be better than that on angular velocity, and the distance normalization could improve the results~\cite{Trung12icb}.

In recent years, due to the rapid development of mobile devices, the accelerometer and gyroscope have been commonly integrated into the smartphones~\cite{WuYZCW19} and smartwatches~\cite{johnston2015smartwatch}. It has been possible to use smartphone for gait recognition~\cite{XuCMU12btas,fernandez2017optimizing,yao2017deepsense} in a wide range of scenarios, e.g, person authentication~\cite{Kwapisz10icb,sun2014gait,le2015smartphone}, medical analysis~\cite{abujrida2017smartphone,juen2014health,ren2015user}, and impersonation-attack defense~\cite{muaaz2017smartphone}. In~\cite{Kwapisz10icb}, data were collected on 36 subjects by putting smartphone in the front pocket of the pants. The activities of walk, jog, climb-up stairs, and climb-down stairs were included for identification and authentication. In~\cite{sun2014gait}, a weighted voting scheme dependent upon the gait characteristic was proposed. The gait characteristic was modeled on the gait frequency, symmetry coefficient, dynamic range and similarity coefficient. In~\cite{le2015smartphone}, gait data were captured by the accelerometer data of smartphones, and the dynamic time warping was employed for gait curve matching. In~\cite{yao2017deepsense}, a deep-learning framework is designed to address the noise and feature customization challenges in a unified manner. In~\cite{juen2014health}, smartphones were used as health monitors, in which the body motion was predicted based on eight parameters of the phone motion in a gait model. In~\cite{ferreira2017user}, a user-centric coordinate system was proposed to represent gait data, and better results were obtained in gait recognition.
\subsection{Deep Learning for Gait Recognition}

Gait biometrics have been widely studied for authentication and access control~\cite{lu2014human,sprager2015efficient,chattopadhyay2014frontal}. In recent years, deep learning has achieved great success in the field of secure computing~\cite{RenWWQL19,zou2020tifs} and activity recognition~\cite{ji20133d}. Deep learning-based gait recognition methods also demonstrated boosted performance over the traditional machine learning-based methods, e.g., SVMs~\cite{man2006individual,lee2002gait,nakajima2003full}.

Various deep neural networks can be used to extract the motion characteristics from the image sequences or the inertial time series. Considering the outstanding ability of CNN in image-feature abstraction, many researchers employed CNNs for gait or activity recognition~\cite{wu2017comprehensive,zeng2014convolutional,ijjina2014one,alotaibi2017improved,takemura2017csvt,yuan2018gait,zhao2017wearable}.  In~\cite{wu2017comprehensive}, three deep CNNs were constructed for gait recognition, using the users' gait energy images as input. Feature maps at different convolutional stages were fused to improve the classification accuracy. In~\cite{takemura2017csvt}, deep CNNs with contrastive loss and triplet ranking loss were proposed for cross-view gait recognition, and high performances were obtained in person authentication and identification. In~\cite{yuan2018gait}, based on the gait data extracted by a periodogram-based gait separation algorithm, deep CNNs were constructed for gait classification.

CNN can also be combined with traditional feature-selection and machine-learning methods such as PCA~\cite{gadaleta2018idnet}, Bayesian classifier~\cite{li2017deepgait} and SVM~\cite{castro2017automatic}. In~\cite{castro2017automatic}, CNN was used as a feature extractor, then the extracted features were classified by SVM. In~\cite{wolf2016multi}, CNN was used to process the three-dimensional data consisting of images and optical flow information for gait and activity recognition.

Meanwhile, the temporal information of human activities is also an important clue for identification. CNNs that only process the single time-stamp data tend to ignore this clue. To address this problem, in~\cite{ji20133d}, a sequence of 2D images were integrated into 3D data, and 3D convolution kernels were employed to extract features for activity recognition. To some extent, the temporal characteristics can be taken into account when CNNs extract the spatial features.

It is more straightforward to introduce RNNs for temporal-information processing. LSTM~\cite{hochreiter1997long} is a typical RNN with specially designed hidden layers that can more effectively extract temporal features from time series. In~\cite{donahue2015long}, LSTM and CNN were combined for activity identification. It is common to use a CNN to extract features from each time-stamp data, and then use an LSTM to further process the entire time-stamp features~\cite{ordonez2016deep,wu2016convolutional}. Extensive comparisons of CNNs and RNNs in the context of gait recognition have been introduced in~\cite{hammerla2016deep}.

In addition to CNN and LSTM, Yu {\it et al.} employed a generative adversarial network (GAN) to construct a gait feature extractor, with the aim of reducing the influence of view angle, weight, and clothing~\cite{yu2017gaitgan}. Sometimes, a gait dataset may contain many subjects; however, each subject may only have a small amount of data, which may not support the training of deep models. To address this problem, in~\cite{zhang2016siamese}, methods based on Siamese neural network were exploited.

The existing researches and results have demonstrated the effectiveness of deep learning in gait and behavior recognition. However, data in these works are mostly collected under limited road conditions with specified walking speed. The identification of gait in unconstrained conditions or free environments remains challenging; and the combination of CNN and LSTM for spatial and temporal feature extraction requires further investigations.

\section{Gait Data Collection in the Wild} \label{sec:datacollect}
In this section, we introduce how to collect gait data using smartphones in the wild. First, we introduce the inertial sensors in smartphones. Then, we describe the algorithm that partitions the inertial data into walking and nonwalking sessions. Finally, we elaborate the segmentation of gait cycles.

\subsection{Inertial Sensors in Smartphones}
Accelerometers and gyroscopes are typical inertial sensors that are equipped by most smartphones.  Accelerometers and gyroscopes measure inertial dynamics in three directions, namely, along the X, Y and Z axes. The three-axis accelerometer is based on the basic principle of acceleration and is used to measure a smartphone's acceleration (including gravity) in the X, Y and Z directions. The accelerations in the three directions reflect the changes in the smartphone's linear velocity in 3D space and, hence, reflect the movement of smartphone users. The three-axis gyroscope captures the angular velocity of a smartphone during its rotation in space, which can also be used to describe the movement pattern of a user. The smartphones used in our work are manufactured by Samsung, Xiaomi and Huawei, all of which run the Android operating system.  When a user is walking, the smartphone accelerates and rotates according to the movements of the user. These data are assumed to be individually unique, so we can collect them as the source data for gait dynamics. The smartphone itself can provide hardware synchronization for the accelerometer and the gyroscope. When the inertial data is recorded through the Android interface, the synchronization may be slightly affected. However, this is tolerable and only has little effect on our gait recognition method.

\begin{figure}[!t]
    \centering
    \includegraphics[width=8cm]{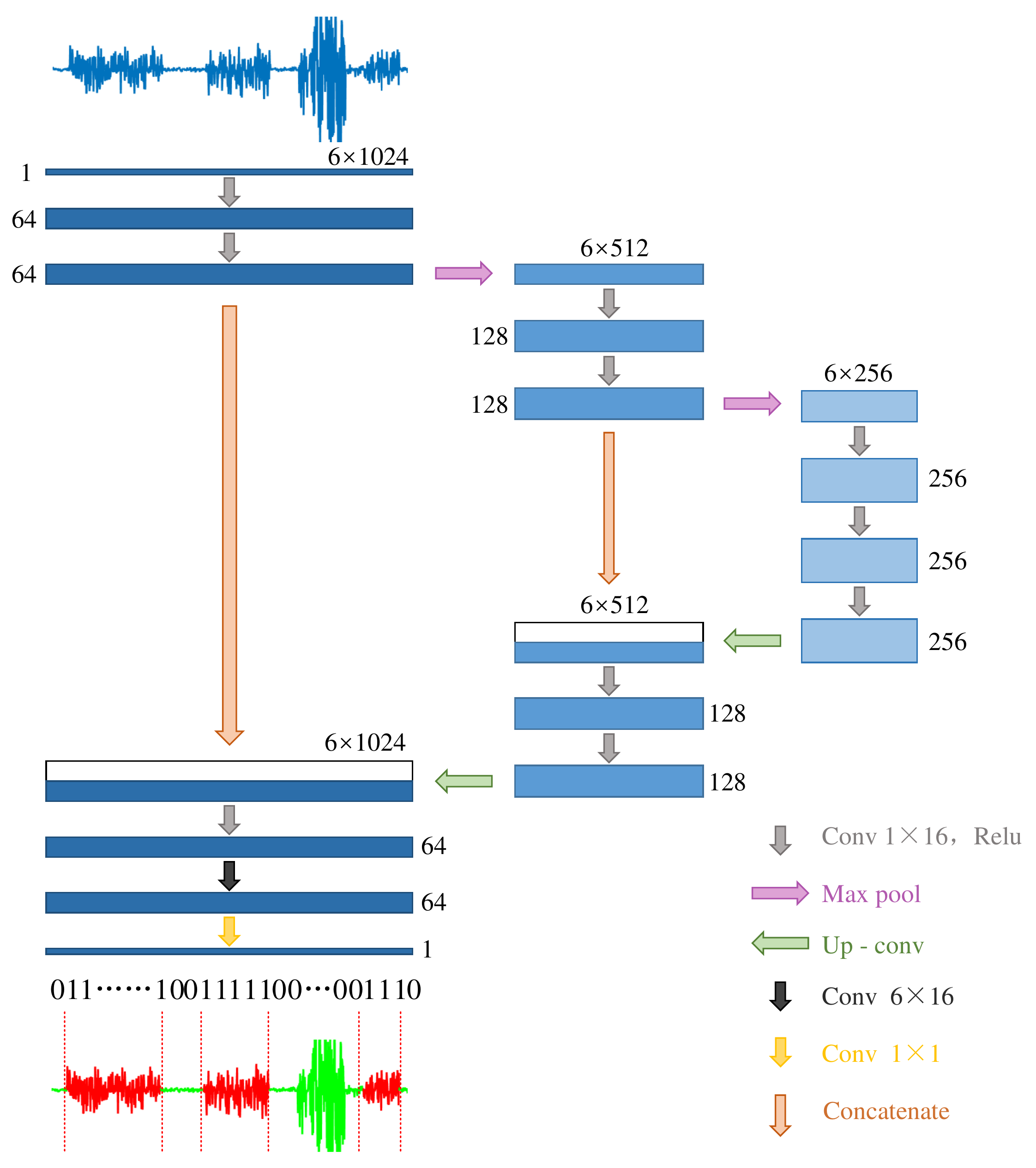}
    \caption{Architecture of the network for gait data extraction.}
    \label{fig:net-seg}
\end{figure}

\subsection{Gait Data Extraction}\label{sec:gaitseg}
When using smartphones to collect the inertial data in the wild, we do not know when, where and how the smartphones will be used. Consequently, the captured data consist of walking sessions and nonwalking sessions, but only walking data are of interest to gait feature extraction and person identification. Thus, the continuous inertial sequence collected by smartphones in the wild must be partitioned.

Considering that walking data and nonwalking data are semantically different and that inertial time series are continuous in both the space and time domains, we model the partitioning problem as a time-series segmentation problem. Inspired by U-Net~\cite{unet2015miccai}, we build a semantic segmentation algorithm with a one-dimensional DCNN. The architecture of the proposed network is shown in Figure~\ref{fig:net-seg}, and the details are listed in Table~\ref{tab:walking and non-walking}. In order to improve the segmentation accuracy, we fuse hierarchical convolutional features from multiple stages in the network.

\subsection{Gait Cycle Segmentation}
As observed in the previous work, cycle-partitioned gait data can often lead to improved performance over non-cycle-partitioned data~\cite{zou2017robust}. This characteristic is validated in our work, which will be introduced in section~\ref{sec:experiment}. Therefore, it is necessary to segment the extracted gait data by walking steps.

\begin{table}[!t]
  \centering
  \caption{Details of the data extraction network structure} \vspace{-2mm}
    \setlength{\tabcolsep}{0.5mm}
    \begin{tabular}{|c|c|c|c|c|}
    \hlinew{1.5pt}
    \textbf{Layer Name} & \textbf{Kernel Size} & \textbf{Kernel Num.} & \textbf{Stride} &\textbf{Feature Map} \\
    \hlinew{1.5pt}
   \hline
    \hline
    conv1$\_$1 & 1$\times$16 & 64 & 1 & 6$\times$1024$\times$64\\
    \hline
    conv1$\_$2 & 1$\times$16 & 64 & 1 & 6$\times$1024$\times$64\\
    \hline
    pool1 & 1$\times$2 & / & 2 & 6$\times$512$\times$64\\
    \hline
    conv2$\_$1 & 1$\times$16 & 128 & 1 & 6$\times$512$\times$128\\
    \hline
   conv2$\_$2 & 1$\times$16 & 128 & 1 & 6$\times$512$\times$128\\
    \hline
    pool2 & 1$\times$2 & / & 2 & 6$\times$256$\times$128\\
    \hline
    conv3$\_$1 & 1$\times$16 & 256 & 1 & 6$\times$256$\times$256\\
    \hline
    conv3$\_$2 & 1$\times$16 & 256 & 1 & 6$\times$256$\times$256\\
    \hline
    conv3$\_$3 & 1$\times$16 & 256 & 1 & 6$\times$256$\times$256\\
    \hline
    upconv1 & 1$\times$2 & 128 & 1 & 6$\times$512$\times$128\\
    \hline
    concat1 & / & / & / & 6$\times$512$\times$256\\
    \hline
    conv4$\_$1 & 1$\times$16 & 128 & 1 & 6$\times$512$\times$128\\
    \hline
    conv4$\_$2 & 1$\times$16 & 128 & 1 & 6$\times$512$\times$128\\
    \hline
    upconv2 & 1$\times$2 & 64 & 1 & 6$\times$1024$\times$64\\
    \hline
    concat2 & / & / & / & 6$\times$1024$\times$128\\
    \hline
    conv5$\_$1 & 1$\times$16 & 64 & 1 & 6$\times$1024$\times$64\\
    \hline
    conv5$\_$2 & 6$\times$16 & 64 & 1 & 1$\times$1024$\times$64\\
    \hline
    conv5$\_$3 & 1$\times$1 & 1 & 1 & 1$\times$1024$\times$1\\
    \hlinew{1.5pt}
    \end{tabular}%
  \label{tab:walking and non-walking}%
\end{table}%

Once the gait data are extracted by the proposed deep semantic segmentation network, we divide the continuous data into separate steps. Notably, a complete step refers to touching the ground with the same foot twice in succession. Sample gait data are shown in Figure~\ref{fig:data-sample-6}, which includes three accelerometer curves and three gyroscope curves. Without loss of generality, we select accelerometer data as the basis for partitioning. As the smartphone can be moved in random directions by the user, a single axis of acceleration or gyroscope cannot stably reflect the fluctuations for the gait curve. Meanwhile, the absolute acceleration along the axis perpendicular to the ground is the largest among the three values. To remove the influence of the phone's orientation, we process the triaxial acceleration data to obtain $ACC_o$ as the basis for gait cycle segmentation. $ACC_o$ is calculated by
{\small $ACC_o = \sqrt{ACC_x^{2}+ACC_y^{2}+ACC_z^{2}}$},
where $ACC_x$, $ACC_y$ and $ACC_z$ denote the values of acceleration in the X, Y, and Z directions, respectively. Figure~\ref{fig:data-sample-2}(a) shows the $ACC_o$ calculated based on $ACC_x$, $ACC_y$ and $ACC_z$ in Figure~\ref{fig:data-sample-6}.

Finally, we find the step-separation points on the local maximums on the $ACC_o$ curve. By observing part of the data of each subject, we set the threshold for each subject's gait cycle (period length and amplitude extrema) to automatically extract the gait cycle, after which a manual check is performed. Specifically, based on our analysis of human steps, the step-separation points are recognized by the following rules:

\begin{figure*}[!t]
    \begin{minipage}[t]{0.32\linewidth}
    \centering
    \includegraphics[height=2cm,width=5.8cm]{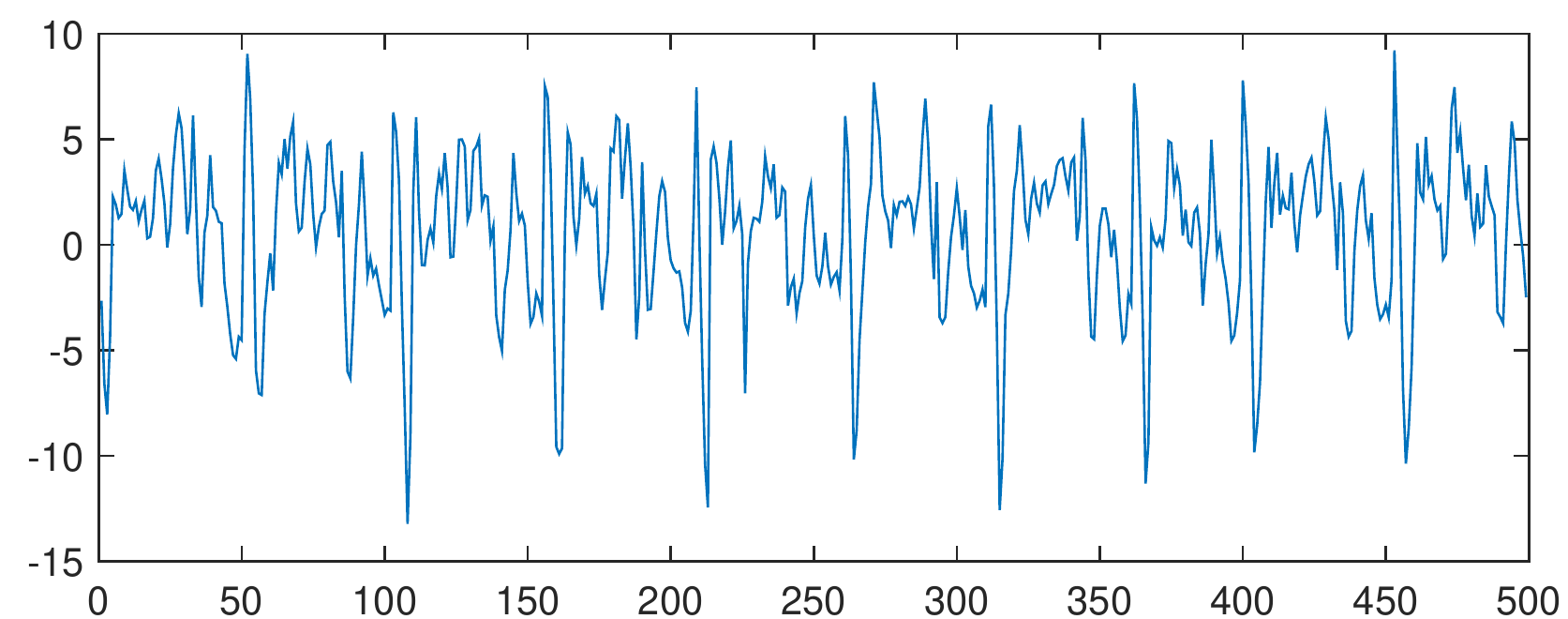}
    \label{fig:side:a}
    \end{minipage}
    \begin{minipage}[t]{0.32\linewidth}
    \centering
    \includegraphics[height=2cm,width=5.8cm]{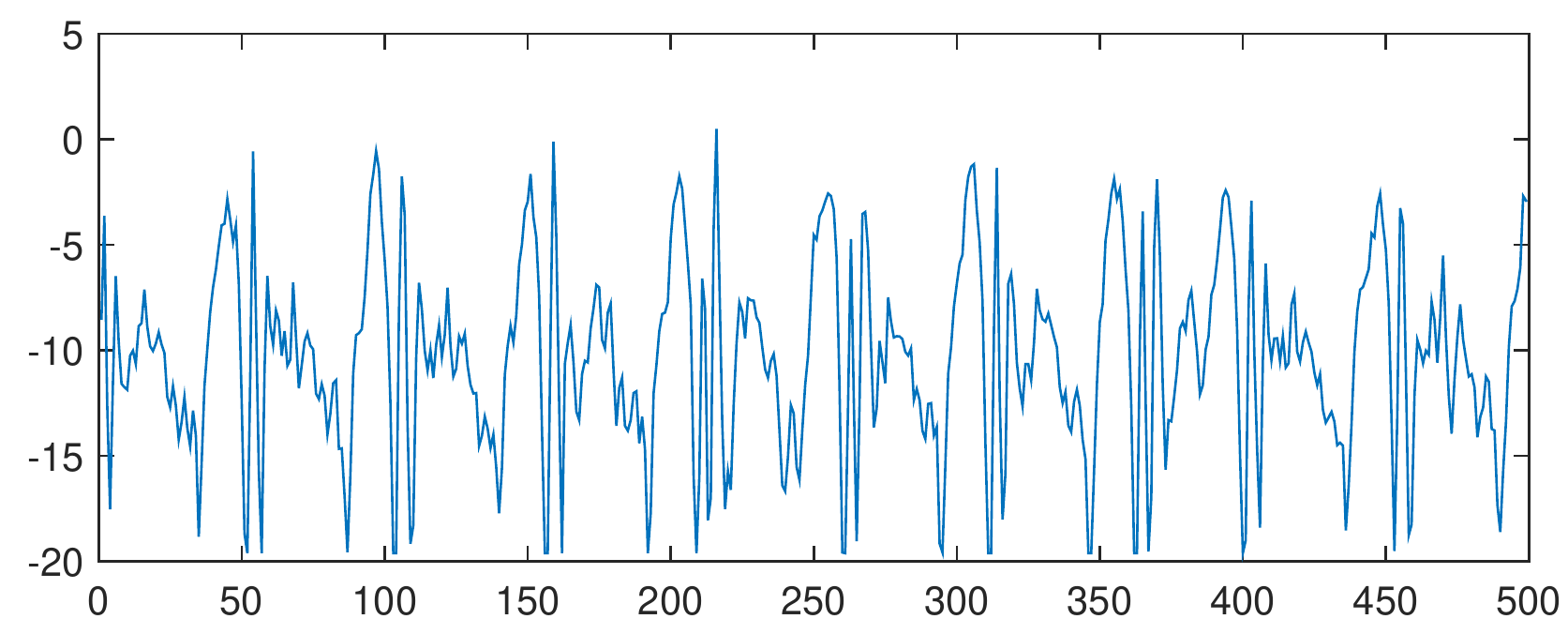}
    \label{fig:side:b}
    \end{minipage}
    \begin{minipage}[t]{0.32\linewidth}
    \centering
    \includegraphics[height=2cm,width=5.8cm]{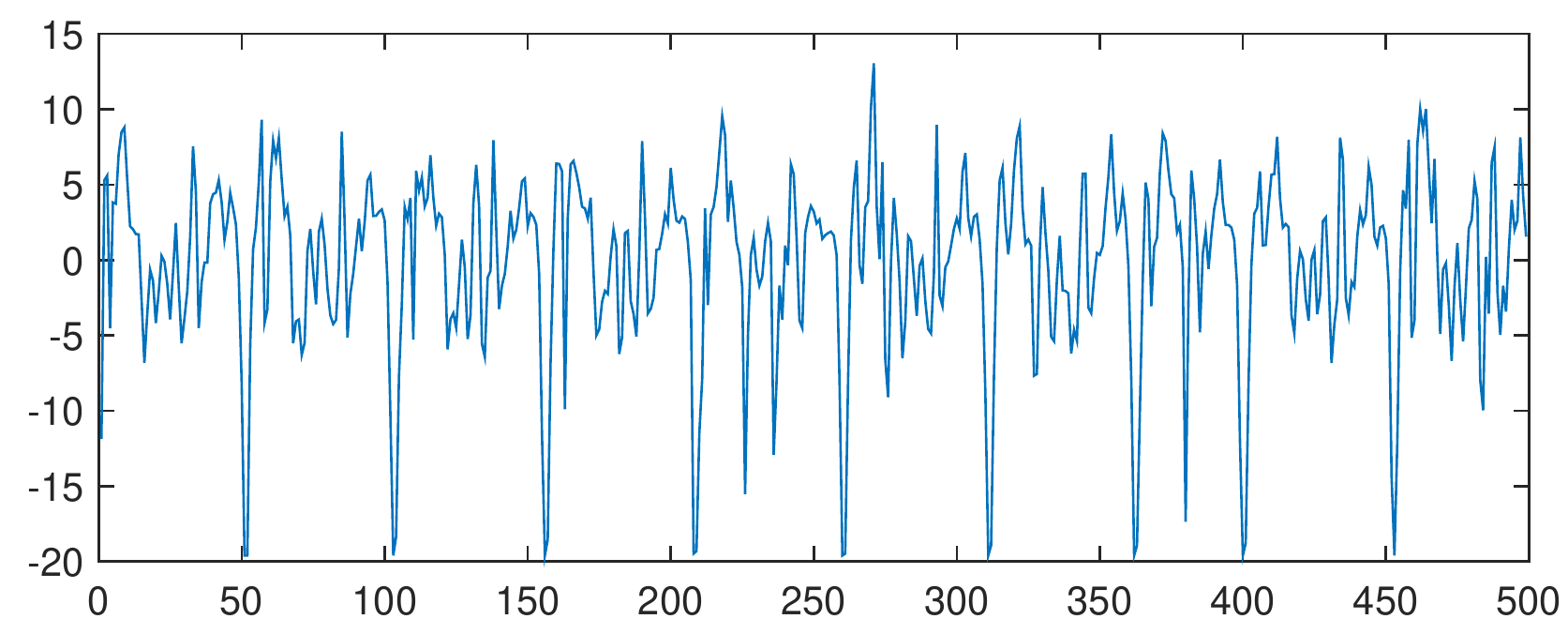}
    \label{fig:side:a}
    \end{minipage}

    \begin{minipage}[t]{0.32\linewidth}
    \centering
    \includegraphics[height=2cm,width=5.8cm]{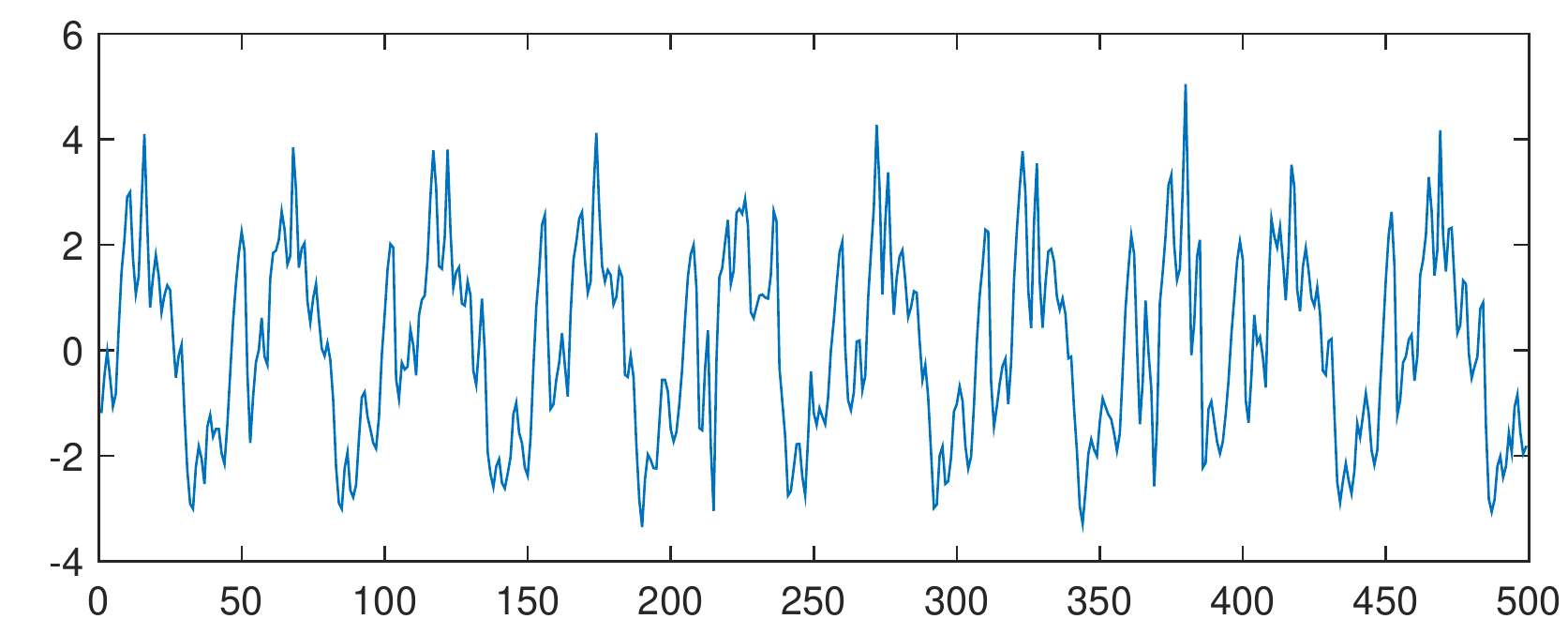}
    \label{fig:side:b}
    \end{minipage}
    \begin{minipage}[t]{0.32\linewidth}
    \centering
    \includegraphics[height=2cm,width=5.8cm]{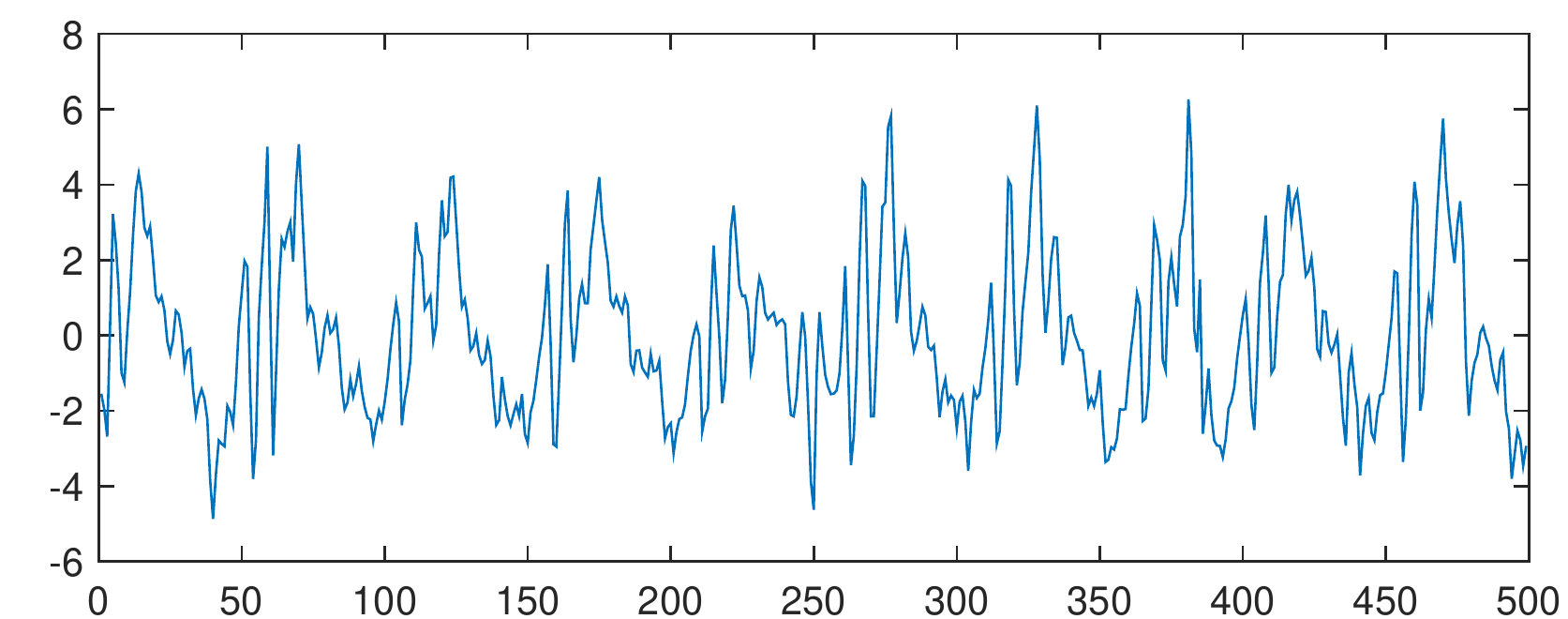}
    \label{fig:side:a}
    \end{minipage}
    \begin{minipage}[t]{0.32\linewidth}
    \centering
    \includegraphics[height=2cm,width=5.8cm]{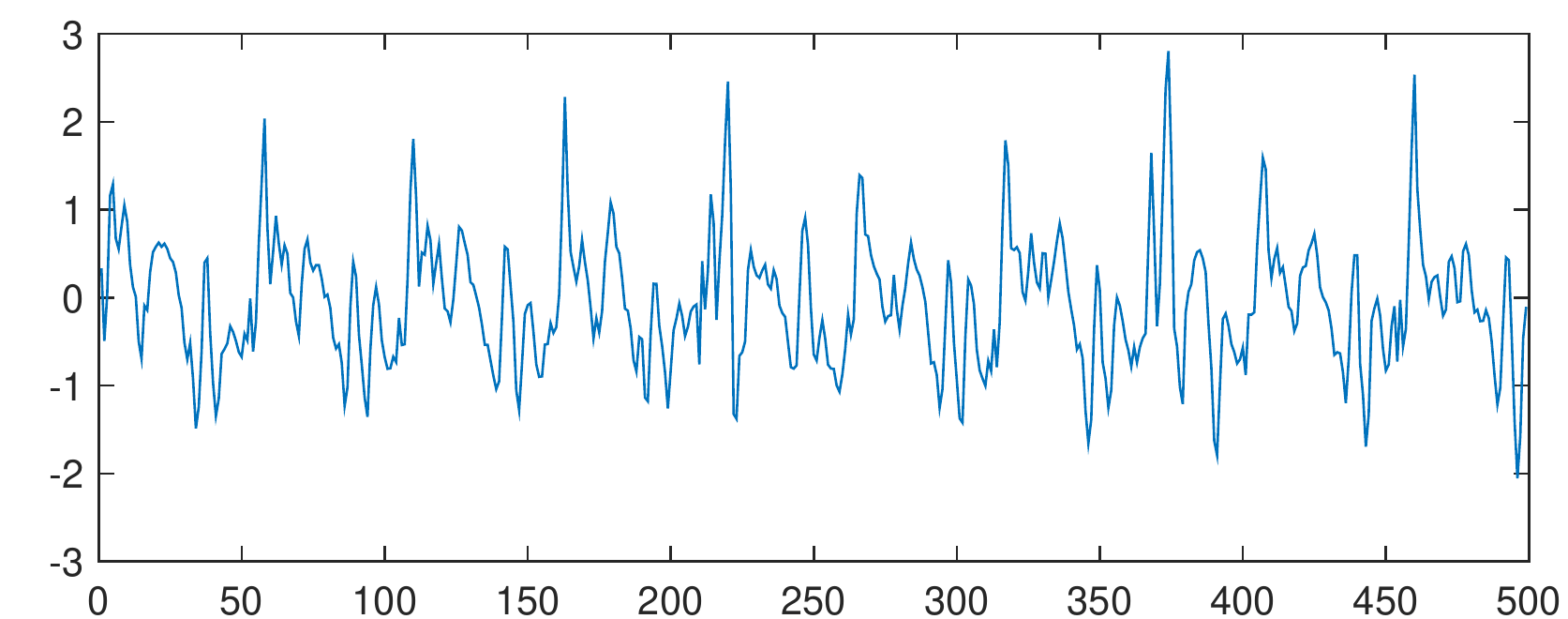}
    \label{fig:data-z}
    \end{minipage}\vspace{-2mm}
    \caption{The inertial data. Top row: $ACC_x$, $ACC_y$, and $ACC_z$, from left to right. Bottom row: $GYR_x$, $GYR_y$, and $GYR_z$, from left to right.}
    \label{fig:data-sample-6}
\end{figure*}

\begin{figure*}[!t]
    \begin{minipage}[t]{0.49\linewidth}
    \centering
    \includegraphics[height=2.6cm,width=8.41cm]{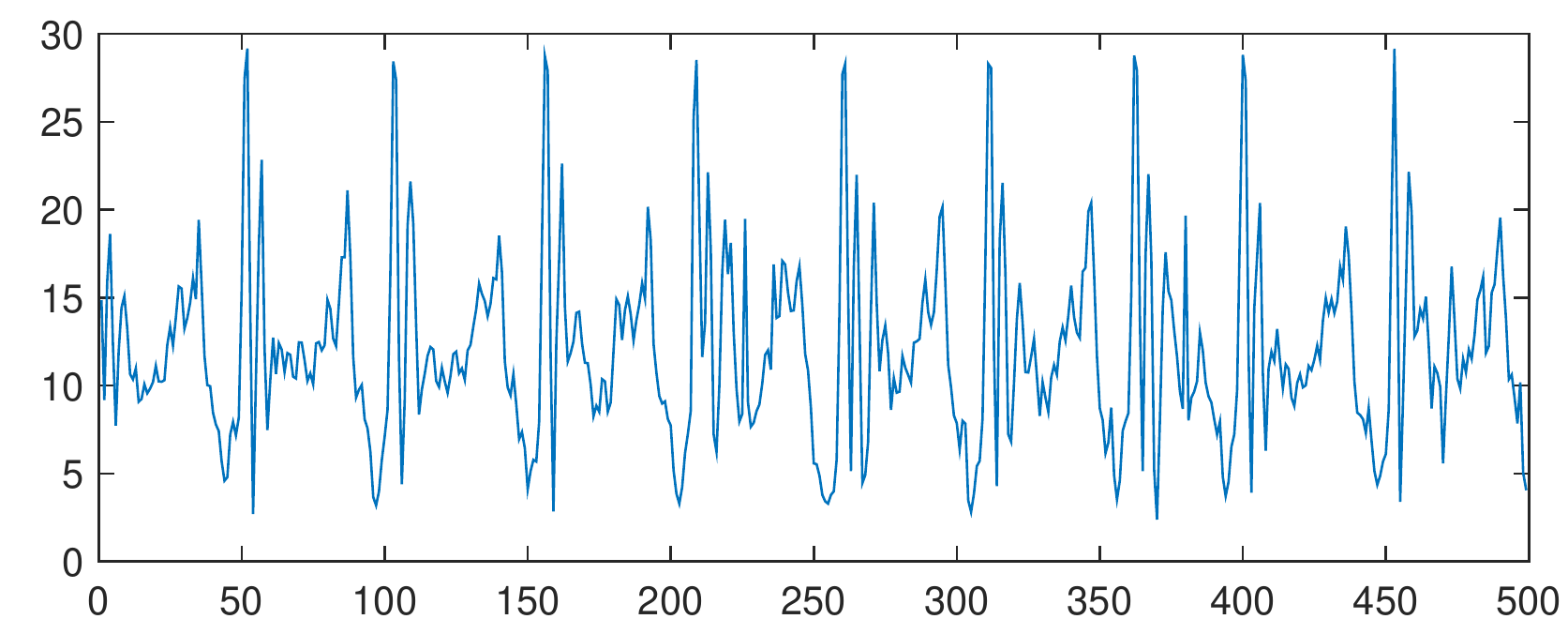}
    \centerline{(a)}
    \label{fig:data-1}
    \end{minipage}\hspace{0mm}
    \begin{minipage}[t]{0.49\linewidth}
    \centering
    \includegraphics[height=2.6cm,width=8.41cm]{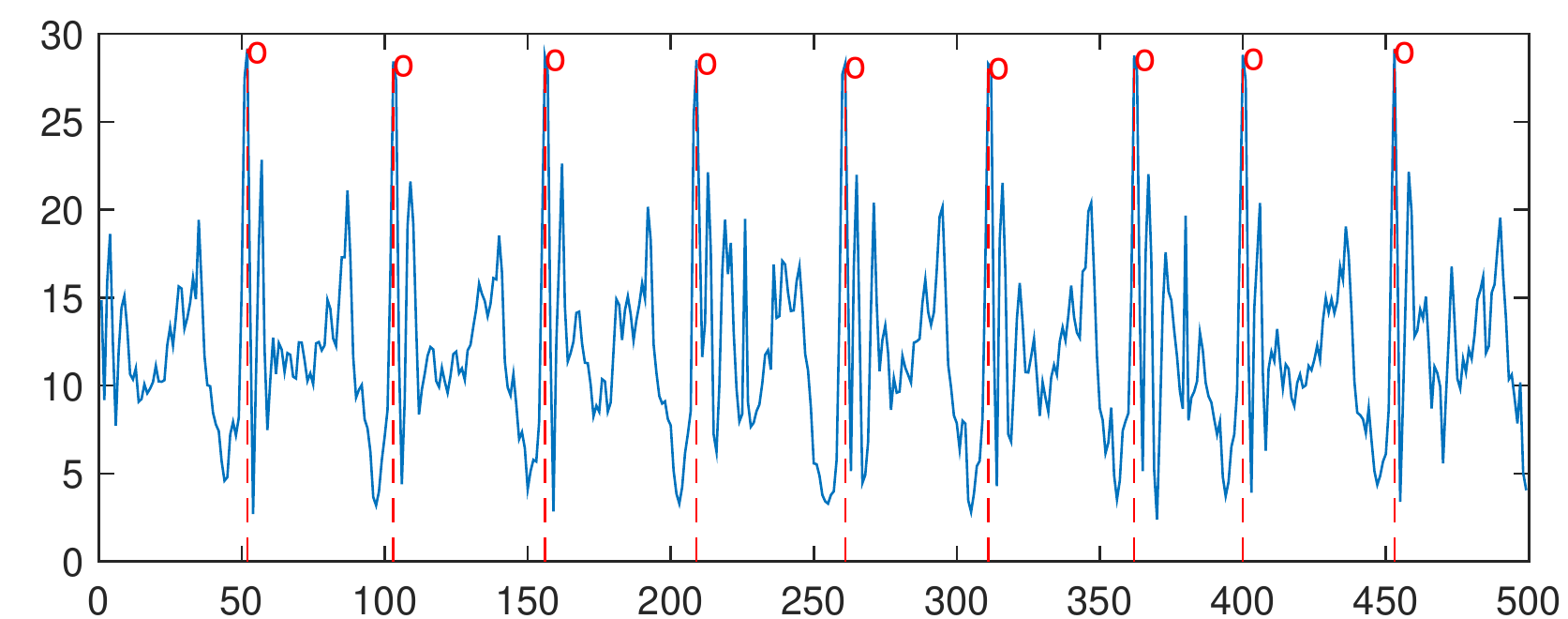}
    \centerline{(b)}
    \label{fig:data-2}
    \end{minipage}\vspace{-3mm}
    \caption{An example of step segmentation. (a) The gait curve $ACC_o$, calculated on the three components as shown in Figure~\ref{fig:data-sample-6}. (b) The segmentation results, where the red dots denote the local maximums and divide the steps.}
    \label{fig:data-sample-2}
\end{figure*}

\begin{itemize}
  \item The point is a local maximum on the $ACC_o$ curve, within a range of 0.8s.
  \item The $ACC_o$ value is larger than 10m/s$^2$ because the acceleration of a falling movement when a person walks should be greater than the gravity acceleration.
  \item The time gap between two consecutive local maxima for step separation should be between 0.8s and 1.6s because it takes approximately 0.8s to 1.6s for a normal person to complete a single step. This is an empirical value obtained by manually examining a small portion of the data from the 118 subjects.
\end{itemize}

\section{Gait Recognition with Deep Neural Networks} \label{sec:method}
In biometrics, gait recognition involves gait identification, that is, identifying a sample from among a given number of candidate identities, and gait authentication, that is, judging if two samples belong to the sample identity. Usually, the former can be modeled as an $n$-class classification problem, while the latter is often modeled as a binary classification problem. As discussed in Section I, we investigate deep learning-based techniques for solving these two problems. Accordingly, we present deep neural networks for gait identification and authentication.

The signals of the accelerometer in axis X, Y and Z at the same time stamp are supposed to be related, and the signals of the accelerometer and gyroscope are also supposed to be related. Under these assumptions, we arrange the tri-axis accelerometer and gyroscope gait data into six-axis inertial gait data. Then, the six-axis gait data are represented as both arrayed signals and time-series signals. As a result, both DCNNs and DRNNs can process these data effectively. To fully utilize their strengths, we employ the DCNN and LSTM respectively to learn the spatial and temporal features, and perform information fusion to improve the feature-representation power.

\begin{figure}[!t]
    \centering
    \includegraphics[width=0.99\linewidth]{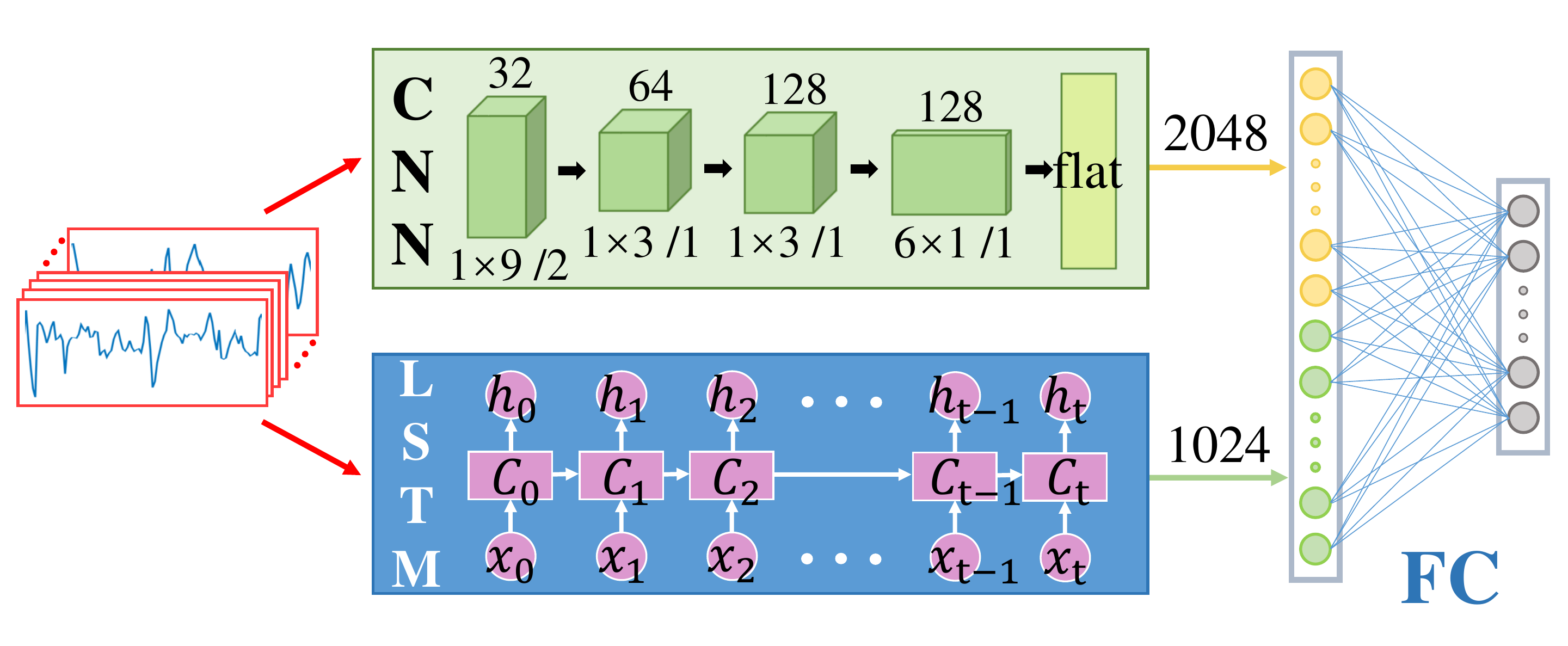}\\
    \caption{The network architecture for gait identification.}
    \label{fig:net-iden}
\end{figure}

Specifically, in our design, the fusion of the DCNN and DRNN is performed in a feature-concatenation manner. The inertial data are processed by a DCNN and an LSTM, independently. The extracted features are then concatenated and fully connected to a successive layer for gait classification. Figure~\ref{fig:net-iden} shows the proposed network architecture.

\subsection{Neural Network for Identification}\label{sec:iden-net}

\subsubsection{\textbf{Problem formulation}}
Given an inertial gait curve $\textbf{x}$ with a sampling length of $T$, $\textbf{x}$ can be expressed as
\begin{equation}\label{eq:vector}
\begin{aligned}
\textbf{x} =(x_1,x_2,...,x_T),\ \ \ \ \ \ \ \ \ \ \ \ \ \ \ \ \\
\mathrm{w.r.t.}, x_t =(acc_x^t,acc_y^t,acc_z^t,gyr_x^t,gyr_y^t,gyr_z^t)^\top,
\end{aligned}
\end{equation}
where ($acc_x^t$, $acc_y^t$, $acc_z^t$) and ($gyr_x^t$, $gyr_y^t$, $gyr_z^t$)  denote the accelerometer and gyroscope components along the X, Y and Z axes at time $t$, respectively. Then, the problem is how to recognize the identity of a subject based on input data $\textbf{x}$. To formulate this problem, suppose $\textbf{s}$=$(s_1,s_2,...,s_n)$ is a number of $n$ candidate subjects, where $s_i$ is the $i$th subject. Then, the output can be represented as an $n$-dimensional vector,
\begin{equation}\label{eq:o}
\begin{aligned}
\mathcal{O}=(o_1,o_2,...,o_n),
\end{aligned}
\end{equation}
where $o_i$=$P(s_i|\textbf{x})$, i.e., the possibility that $\textbf{x}$ belongs to $s_i$. Further suppose that $s$ is the identity of the input data $\textbf{x}$; then, it is formulated as
\begin{equation}\label{eq:yo}
s=arg \max\limits_{s_i}\{o_i\ |\ 1\leq i\leq n\}.
\end{equation}
Thus, to solve the problem of gait identification, we have to associate the maximum possibility values with the corresponding subjects.

\subsubsection{\textbf{Network structure}}

As illustrated in Figure~\ref{fig:net-iden}, the gait-identification network consists of a CNN and an LSTM in parallel. The CNN and LSTM work as two feature extractors to obtain the corresponding features -- $feat_{cnn}$ and $feat_{lstm}$. The CNN and LSTM are followed by a fully connected layer, which works as a classifier and uses the feature vector obtained by concatenating $feat_{cnn}$ and $feat_{lstm}$ as the input.

\emph{LSTM network.}  LSTM is an improvement to RNNs and still has the basic structure of RNN. For an RNN network with $L$ hidden layers, given a gait sequence $\textbf{x}=(x_1,x_2,...,x_T)$, a state $\textbf{h}_t^l$ will be generated for each layer at time $t$:
\begin{equation}\label{eq:lstm-ht}
\begin{aligned}
\textbf{h}_t^l=\sigma(W_{xh}^lx_t+\textbf{h}_{t-1}^lW_{hh}^{tl}+\textbf{h}_t^{l-1}W_{hh}^{ll}+\textbf{b}_h^l),
\end{aligned}
\end{equation}
where $\textbf{h}_t^l$ is the state of layer $l$ at time $t$, $x_t$ is input at time $t$, $W_{xh}^l$ is the weight matrix of the input $x_t$ to the $l$th hidden layer, $W_{hh}^{tl}$ is the weight matrix of the state at time $t$-1 to the state at time $t$ at the same layer $l$, $W_{hh}^{ll}$ is the weight matrix of the state at layer $l$-1 to the state at layer $l$ at the same time $t$, $\textbf{b}_h^l$ is the bias of layer $l$, and $\sigma(\cdot)$ is the activation function.

For an LSTM network, the basic unit is composed of a cell, an input gate, an output gate and a forget gate. Similar to an RNN, we can also examine the LSTM network with a state $\textbf{h}_t^l$. To obtain better memory for information interaction along the time series, an input gate \textbf{i}, a forgetting gate \textbf{f}, a state vector $\textbf{c}$, and an output gate \textbf{o} are added to the hidden layer state. Then, $\textbf{h}_t^l$ can be updated as follows:
\begin{equation}\label{eq:lstm-five}
\begin{aligned}
\textbf{i}_t&=\sigma_i(W_{xi}x_t+W_{hi}^t\textbf{h}_{t-1}^l+W_{hi}^l\textbf{h}_t^{l-1}+W_{ci}\textbf{c}_{t-1}+\textbf{b}_i),\\
\textbf{f}_t&=\sigma_f(W_{xf}x_t+W_{hf}^t\textbf{h}_{t-1}^l+W_{hf}^l\textbf{h}_t^{l-1}+W_{cf}\textbf{c}_{t-1}+\textbf{b}_f),\\
\textbf{c}_t&=\textbf{f}_t+\textbf{c}_{t-1}\sigma_i(W_{xc}x_t+W_{hc}^t\textbf{h}_{t-1}^l+W_{hc}^l\textbf{h}_t^{l-1}+\textbf{b}_c),\\
\textbf{o}_t&=\sigma_o(W_{xo}x_t+W_{ho}^t\textbf{h}_{t-1}^l+W_{ho}^l\textbf{h}_t^{l-1}+W_{co}\textbf{c}_{t-1}+\textbf{b}_o),\\
\textbf{h}_t^l&=\textbf{o}_t\sigma_h(\textbf{c}_t),
\end{aligned}
\end{equation}
where $W$, $\sigma$, $\textbf{i}_t$, $\textbf{f}_t$, $\textbf{c}_t$, and $\textbf{o}_t$ are the parameters of layer $l$, as indicated by the hidden superscript `$l$'. $W_{xi}$ is the weight matrix of the input $x_t$ to the input gate, $\sigma_i$ is the activation function of the input gate, and $\textbf{b}_i$ is the bias of the input gate. The meanings of $W$, $\textbf{b}$, and $\sigma$ can be inferred from the above rule. Given that the LSTM network is constructed with $L$ hidden layers, with each containing $N$ hidden nodes, then, for each input $\textbf{x}=(x_1,x_2,...,x_T)$, the output feature can be formulated as
\begin{equation}\label{eq:lstm-five}
\begin{aligned}
feat_{lstm}&=\textbf{h}_T^L\\
&=(f_1,f_2,...,f_{N}).
\end{aligned}
\end{equation}

\begin{table}[!t]
  \centering
  \caption{Details of the CNN structure} \vspace{-2mm}
    \setlength{\tabcolsep}{0.5mm}
    \begin{tabular}{|c|c|c|c|c|}
    \hlinew{1.5pt}
    \textbf{Layer Name} & \textbf{Kernel Size} & \textbf{Kernel Num.} & \textbf{Stride} &\textbf{Feature Map} \\
    \hlinew{1.5pt}
    \hline
    \hline
    conv1$\_$1 & 1$\times$9 & 32 & 2 & 6$\times$64$\times$32\\
    \hline
    pool1 & 1$\times$2 & / & 2 & 6$\times$32$\times$32\\
    \hline
    conv2$\_$1 & 1$\times$3 & 64 & 1 & 6$\times$32$\times$64\\
    \hline
    conv2$\_$ 2 & 1$\times$3 & 128  & 1 & 6$\times$32$\times$128 \\
    \hline
    pool2 & 1$\times$2 & / & 2 & 6$\times$16$\times$128\\
    \hline
    conv3$\_$1 & 6$\times$1 & 128 & 1 & 1$\times$16$\times$128\\
    \hlinew{1.5pt}
    \end{tabular}%
  \label{tab:cnnarchi}%
\end{table}%

\emph{CNN network.} Considering that the input signals are time series, one-dimensional kernels are used in the convolution operations in the proposed CNN network. Specifically, the proposed CNN network is constructed with 4 convolutional layers and 2 max-pooling layers. The convolution kernel abstracts the feature of the gait curve along the time series, and the max-pooling operation downsamples the feature map in the pooling window. Table~\ref{tab:cnnarchi} describes the detailed structure of the proposed CNN network. The output CNN feature has dimensions of 1$\times$16$\times$128. Then, we flatten the output convolutional feature map into a one-dimensional feature vector $feat_{cnn}$.

\begin{figure}[!t]
    \centering
    \includegraphics[width=0.98\linewidth]{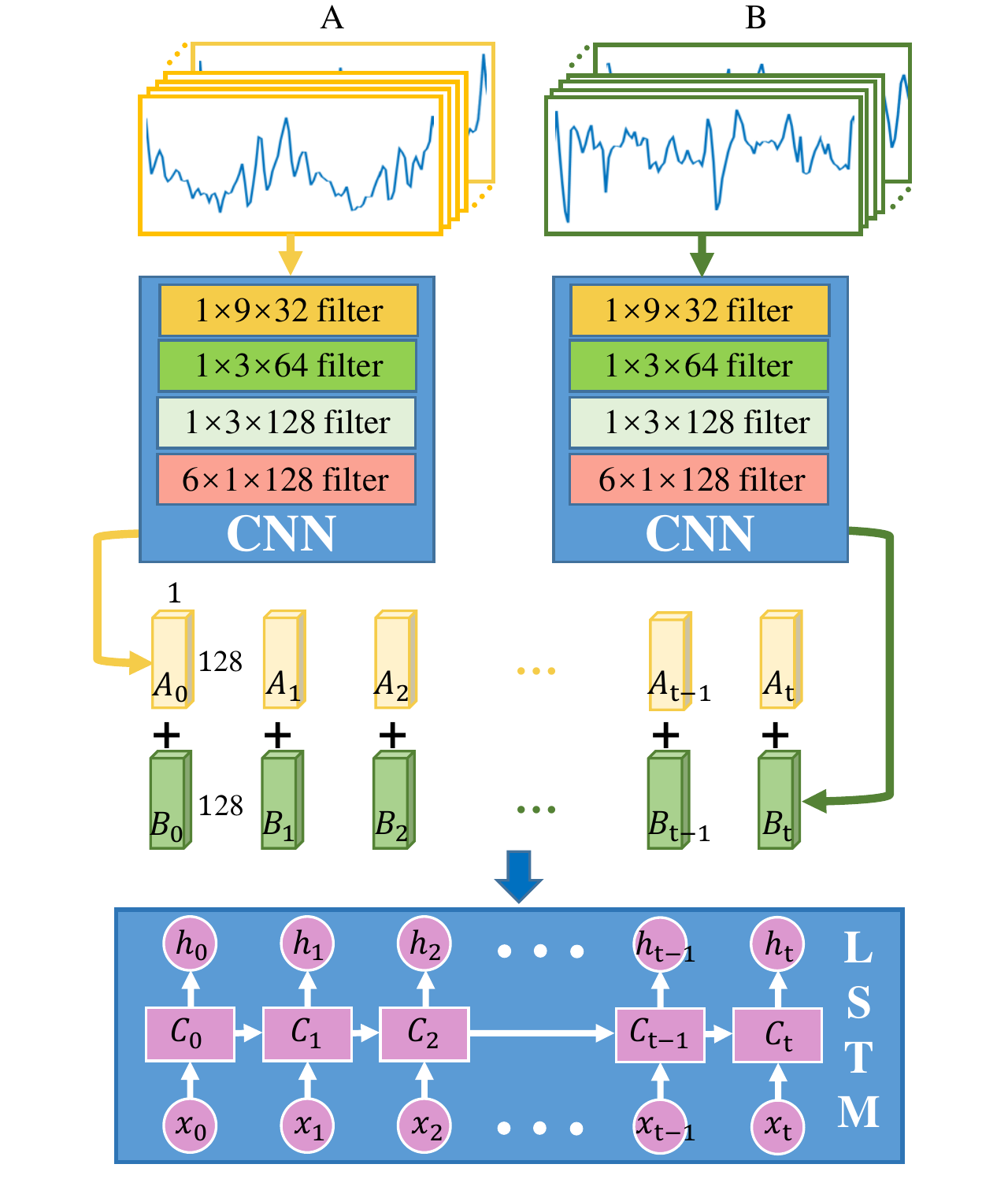}\\
    \caption{The network architecture for gait authentication.}
    \label{fig:net-auth}
\end{figure}

\emph{Fully Connected Layer.} The fully connected layer is the concatenation of the features extracted by LSTM and CNN, i.e., $feat_{full}$=$(feat_{lstm}; feat_{cnn})$. Then, a softmax operation is applied to produce the classification output, as formulated by Eq.~(\ref{eq:lstm-soft}),
\begin{equation}\label{eq:lstm-soft}
\textbf{o} = \mathbf{Softmax}(feat_{full} * W_o + \textbf{b}_o),
\end{equation}
where $W_o$ is the weight matrix of the output layer, and $\textbf{b}_o$ is the bias of the output layer.

\subsubsection{\textbf{Loss Function}}

For each input sample $\textbf{x}$, the predicted output of the network is $\textbf{o}=(o_1, o_2, ..., o_n)$, with $o_i=P(s_i|\textbf{x})$. The value of $o_i$ is between 0 and 1. The larger the value is, the greater the probability that \textbf{x} belongs to $s_i$ is. Based on the output $\textbf{o}$, we can obtain the class label as
\begin{equation}\label{eq:lstm-classlabel}
\textbf{o}' = (o'_1, o'_2, ..., o'_n),
\end{equation}
where
\begin{equation}\label{eq:lstm-case}
o'_i=
\begin{cases}
1& \textbf{x}\in s_i,\\
0& other.
\end{cases}
\end{equation}
Then, we can construct the training loss via cross entropy, as formulated by Eq.~(\ref{eq:lstm-loss}),
\begin{equation}\label{eq:lstm-loss}
\mathcal{L}(\textbf{o},\textbf{o}')=\sum_i^no'_i\ln o_i+(1-o'_i)\ln(1-o_i).
\end{equation}
As shown by Eq.~(\ref{eq:lstm-loss}), the cross entropy is a positive number. When $o_i \approx 0, o'_i = 0$ or $o_i \approx 1, o'_i = 1$, the cross entropy is small. In other words, a larger difference between $o_i$ and $o'_i$ results in a larger cross entropy. This property helps the convergence of the network during training.

\subsection{Neural Network for Authentication}

Let two sequences of gait data $\textbf{x}_\textbf{a}$ and $\textbf{x}_\textbf{b}$ be the input of the authentication network, which are expressed as
\begin{equation}\label{eq:auth-input}
\begin{aligned}
\textbf{x}_\textbf{a}&=(x_{a,1},x_{a,2},...,x_{a,T}), \\
\textbf{x}_\textbf{b}&=(x_{b,1},x_{b,2},...,x_{b,T}), \\
\end{aligned}
\end{equation}
where
\begin{equation}\label{eq:auth-xt}
x_t=(ACC_x^t,ACC_y^t,ACC_z^t,GYR_x^t,GYR_y^t,GYR_z^t)^\top,
\end{equation}
and $T$ is the length of the input sequence. As discussed in the beginning of Section~\ref{sec:method}, authentication is formulated as a binary classification problem. The output of the network is set as two dimensions. We use `True' and `False' to denote that the input data are from the same subject and different subjects, respectively.

\begin{table*}[ht]
  \centering   \small
  \caption{Detail information of the whuGAIT datasets.} \vspace{-1mm}
    \begin{tabular}{|c|c|c|c|c|c|c|c|}
    \hlinew{1.5pt}
    \textbf{Dataset Name} & \tabincell{c}{\textbf{Usage}} &\tabincell{c}{\textbf{Number}\\ \textbf{of Subjects}} & \tabincell{c}{\textbf{Time-fixed} \\\textbf{or Interpolation}} & \tabincell{c}{\textbf{Overlap}\\\textbf{in Sampling}} & \tabincell{c}{\textbf{Samples}\\\textbf{for Training}}& \tabincell{c}{\textbf{Samples} \\\textbf{for Test}} & \textbf{Alignment} \\
    \hlinew{1.5pt}
    \hline
    \hline \renewcommand\arraystretch{1.4}
    \tabincell{c}{Dataset \#1} & \tabincell{c}{ Classification } & \tabincell{c}{ 118 } & \tabincell{c}{Interpolation}  & \tabincell{c}{1 step} & \tabincell{c}{33,104} & \tabincell{c}{3,740} & \tabincell{c}{N/A}     \\
    \hline \renewcommand\arraystretch{1.4}
    \tabincell{c}{Dataset \#2} & \tabincell{c}{ Classification } & \tabincell{c}{ 20 } & \tabincell{c}{Interpolation}  & \tabincell{c}{0} & \tabincell{c}{44,339} & \tabincell{c}{4,936} &  \tabincell{c}{N/A}    \\
    \hline \renewcommand\arraystretch{1.4}
    \tabincell{c}{Dataset \#3} & \tabincell{c}{ Classification }   & \tabincell{c}{118} &  Time-fixed  & \tabincell{c}{1 step} & 26,283 & 2,991 &  \tabincell{c}{N/A}    \\
    \hline \renewcommand\arraystretch{1.4}
    \tabincell{c}{Dataset \#4 }& \tabincell{c}{ Classification }   & \tabincell{c}{20} &  Time-fixed  & \tabincell{c}{0} & \tabincell{c}{35,373} & \tabincell{c}{3,941} &  \tabincell{c}{N/A}    \\
    \hline \renewcommand\arraystretch{1.4}
    \tabincell{c}{Dataset \#5} & \tabincell{c}{ Authentication }   & 118   &  Interpolation  & 1 step   & 66,542 & 7,600  & Horizontal    \\
    \hline \renewcommand\arraystretch{1.4}
    \tabincell{c}{Dataset \#6} & \tabincell{c}{ Authentication }   & 118   &  Interpolation  & 1 step   & 66,542 & 7,600  & Vertical    \\
    \hlinew{1.5pt}
    \end{tabular} \label{tab:datasets} \\ \vspace{1mm}
  *Note: there is no overlap between the training sample and the test sample for all datasets.
\end{table*}%

To fully utilize the advantages of the CNN and RNN, we use the CNN as a feature extractor to map the input inertial signals into lower-dimensional abstractions. In our design, we use the CNN that is trained on the dataset from 98 subjects in the classification in Section~\ref{sec:iden-net} as the feature extractor. These 98 subjects have no overlap with the 20 subjects for the test in the authentication. Figure~\ref{fig:net-auth} shows the structure of the authentication network, where the CNN is fixed as a feature extractor. Given that the size of the input gait signal is 6$\times$128, the output of the CNN is 1$\times$16$\times$128. For the subsequent LSTM computation, we must ensure that the input signals have the properties of time series. Therefore, we rearrange the CNN features into 16$\times$256 for a pair of inputs, which are divided into 16 blocks, with each containing a 256-dimensional feature vector. The 16 blocks of features are then fed into a double-layer LSTM for training and prediction. For the CNN network, the weights are fixed to those obtained by training in the identification network.

\section{Experiments and Results} \label{sec:experiment}
In this section, we first introduce six datasets that contain inertial gait data collected using smartphones in the wild. Then, we will describe the experimental settings, including the way of data alignment, the selection of comparison methods, and the strategy of training, etc. Finally, we will report the evaluation results for both the identification case and the authentication case.

\subsection{{\textbf{whuGAIT}} Datasets} \label{sec:datasets}
Generally, deep-learning methods require a large number of samples for training, and existing datasets cannot meet the demand. Meanwhile, there are few datasets collecting unconstrained inertial data in free environments for gait recognition. We collect the inertial gait data in the wild, where the subjects are not limited to walking on specific roads or speeds. Data are collected in daily life, such as a walk after a meal. Note that, all the collected data have been pre-processed by the gait-extraction algorithm introduced in Section~\ref{sec:gaitseg}. A number of 118 subjects are involved in the data collection. Among them, 20 subjects collect a larger amount of data in two days, with each holding thousands of samples, and 98 subjects collect a smaller amount of data in one day, with each holding hundreds of samples. Each data sample contains the 3-axis acceleromter data and the 3-axis gyroscope data. The sampling rate of all sensor data is 50Hz. According to different evaluation purposes, we construct six datasets based on the collected data.
\subsubsection{Dataset \#1}
This dataset is collected on 118 subjects. Based on the step-segmentation algorithm introduced in Section~\ref{sec:gaitseg}, the collected gait data can be annotated into steps. Following the findings that two-step data have a good performance in gait recognition~\cite{zou2017robust}, we collected gait samples by dividing the gait curve into two continuous steps. Meanwhile, we interpolate a single sample into a fixed length of 128 using the linear interpolation function. In order to enlarge the scale of the dataset, we make a one-step overlap between two neighboring samples for all subjects. In this way, a total number of 36,844 gait samples are collected. These samples are sorted by time. For each subject, we select the first 90\% samples for training and the rest 10\% for test. There are 33,104 training samples and 3,740 test samples, without overlap between the two subsets.

\subsubsection{Dataset \#2}
This dataset is collected on 20 subjects.  We also divide the gait curve into two-step samples and interpolate them into the same length of 128. As each subject in this dataset has a much larger amount of data as compared to that in Dataset \#1, we do not make overlap between the samples. Finally,  a total number of of 49,275 samples are collected, in which 44,339 samples are used for training, and the rest 4,936 for test.

\subsubsection{Dataset \#3}
This dataset is collected on the same 118 subjects as in Dataset \#1. Different from Dataset \#1, we divide the gait curve by using a fixed time length, instead of a step length. Exactly, we collect samples with a time interval of 2.56 seconds. While the frequency of data collection is 50Hz, the length of each sample is also 128. Also, we make an overlap of 1.28 seconds to enlarge the dataset. A total number of 29,274 samples are collected, in which 26,283 samples are used for training, and the rest 2,991 for test.

\subsubsection{Dataset \#4} This dataset is collected on 20 subjects.  We also divide the gait curve in an interval of 2.56 seconds. We make no overlap between the samples. Finally,  a total number of 39,314 samples are collected, in which 35,373 samples are used for training, and the rest 3,941 for test.

\begin{figure*}[!t]
    \centering
    \includegraphics[width=0.95\linewidth]{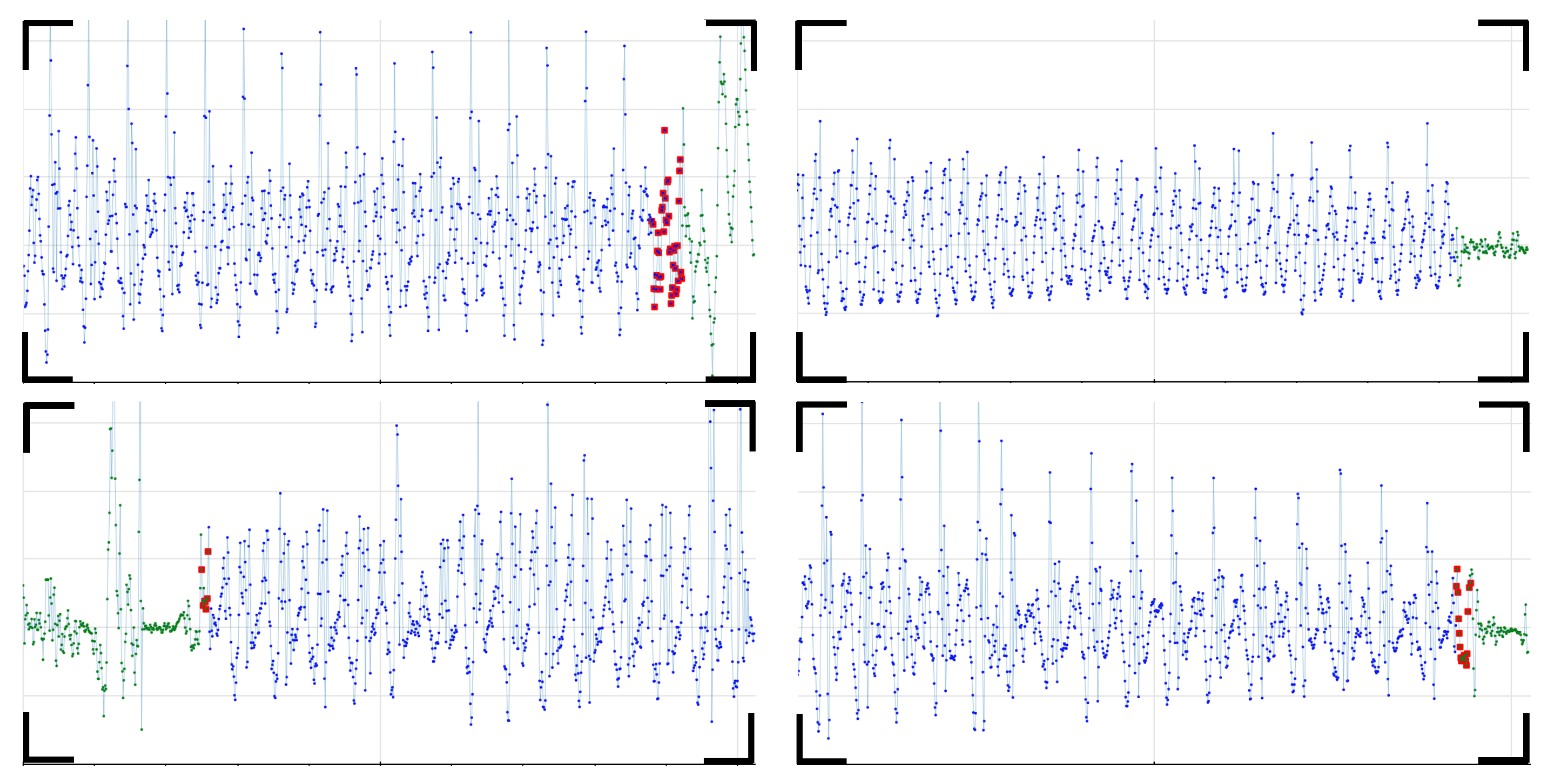}\\
    \caption{Four examples of walking data extraction using the proposed method. Note that, the blue points denote the walking data, green points denote the non-walking data, and the red denotes the false classified.}
    \label{fig:data extraction}
\end{figure*}
\subsubsection{Dataset \#5}
This dataset is used for authentication. It contains 74,142 authentication samples of 118 subjects, where the training set is constructed on 98 subjects and the test set is constructed on the other 20 subjects. There are 66,542 samples and 7,600 samples for training and test, respectively. Positive and negative samples each accounts for half of
the total samples. Each authentication sample contains a pair of data sample that are from two different subjects or one same subject. The data sample consists of a 2-step acceleration and gyroscopic data, which are interpolated in the way as described in Dataset \#1. The two data samples are horizontally aligned to create an authentication sample.
\subsubsection{Dataset \#6}
This dataset is also used for authentication. The authentication samples are constructed the same way as in Dataset \#5. The only difference is that, in authentication sample construction, two data samples from two subjects are vertically aligned instead of horizontally aligned.

Table~\ref{tab:datasets} shows the detail information of these six datasets.

\subsection{Implementation Details}
\subsubsection{Data collection}
We developed an Android APP and installed it on smartphones manufactured by Samsung, Xiaomi and Huawei. The frequency of the accelerometer and the gyroscope is set to 50Hz, and the data from these two sensors are recorded in real time. When using the APP, the user inputs his own identity information and starts the data collection process. The user can place the phone in his/her trouser pocket, hold it in the hand, or place it on a desk without constraints. Notably, the APP must collect data for a long period of time to ensure that the captured data contain sufficient walking data. Meanwhile, the APP automatically deletes the data if the smartphone is static for a period of 3 seconds. The captured data have seven dimensions, namely, the time stamp, the triaxial values of the acceleration sensor and the triaxial values of the gyroscope. We implement our networks and the comparison networks with Tensorflow.

\subsubsection{Identification experiment}
A number of network structures, including LSTM-based, CNN-based, and CNN+LSTM-based, are designed for gait classification, and their performances are compared in terms of accuracy.

For LSTM-based methods, each hidden layer in the LSTM has $N$=64 hidden nodes, the learning rate is set to 0.0025, and the number of epochs for training is 200. For the hybrid method, $N$=1024 is set for LSTM.

For CNN-based methods, the six-axis interpolation data are used as the input, with the data shaped as 6$\times$128. The classification experiments are conducted on the first four datasets introduced in Section~\ref{sec:datasets}. When training the CNN, the learning rate is 0.0025, and the number of epochs for training is 200.

\subsubsection{Authentication experiment}
As has been introduced in Figure~\ref{fig:net-auth}, the authentication network contains a CNN and an LSTM. In the training process, parameters of CNN are frozen, and the LSTM network equipped with 64-node hidden layers is trained with a learning rate of 0.0025, an epoch number of 300 and a batch size of 1,500.

\subsection{Performance on Gait Data Extraction}
\begin{table}[t!]
  \centering   \small
  \caption{Detail information of the gait-data extraction datasets}\vspace{-1mm}
    \begin{tabular}{|c|c|c|c|}
    \hlinew{1.5pt}
    \tabincell{c}{\textbf{Dataset}\\ \textbf{Name}}   &\tabincell{c}{\textbf{Number}\\ \textbf{of Subjects}} &  \tabincell{c}{\textbf{Samples}\\\textbf{for Training}}& \tabincell{c}{\textbf{Samples} \\\textbf{for Test}} \\
    \hlinew{1.5pt}
    \hline
    \hline
    Dataset \#7 & 10  & 519 &58 \\
    \hline
    Dataset \#8 & 118  & 1022 & 332 \\
    \hlinew{1.5pt}
    \end{tabular}%
  \label{tab:data extraction}
\end{table}%

\subsubsection{Datasets}

Two datasets are constructed for evaluation of the proposed gait-data-extraction method. Basic information of the two datasets have been shown in Table~\ref{tab:data extraction}, and the details are given as below:
\begin{enumerate}[   $\vcenter{\hbox{\tiny$\bullet$}}$]
    \item Dataset \#7: it contains 577 samples of 10 subjects, with data shaped as 6$\times$1,024. Among these samples, 519 are used for training and 58 are for test. Both the training and test samples are from the 10 subjects.
    \item Dataset \#8: it contains 1,354 samples of 118 subjects, with data shaped as 6$\times$1,024. Among these data, 1,022 samples from 20 subjects are used for training, and 332 samples from the other 98 subjects are used for test.
\end{enumerate}
For both datasets, each sample is attached with a label file, which contains 1,024 binary values, with `1' as the walking data, and `0' as the non-walking data. The labels are manually annotated.

\subsubsection{Experimental results}

\begin{figure*}[!t]
    \centering
    \includegraphics[width=0.85\linewidth]{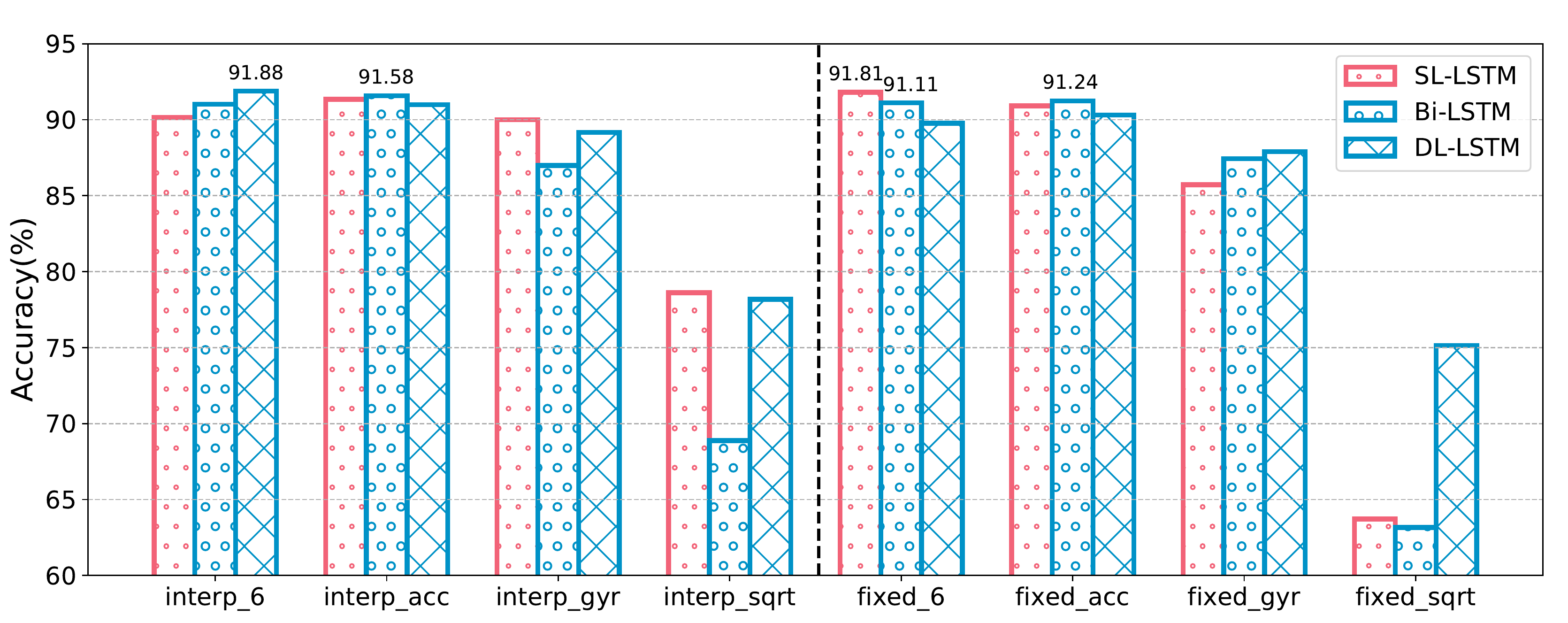} \vspace{-2mm}
    \caption{Performance of different LSTM networks. The experiments are conducted on 118 subjects. For each group of results, the left, middle, and right bars correspond to the results of the single-layer LSTM (SL-LSTM), the bi-directional LSTM (Bi-LSTM) and the double-layer LSTM (DL-LSTM), respectively.}\label{fig:lstms}
\end{figure*}

We train the CNN network proposed in Section~\ref{sec:gaitseg} for gait data extraction. For Dataset \#7 and Dataset \#8, the learning rate is set to 0.0001, and the number of training epochs is set to 150. Figure~\ref{fig:data extraction} shows four sample results obtained by the proposed network. Most of the data are correctly classified: a small portion of walking data are extracted as nonwalking (red on blue) and a small portion of nonwalking data are extracted as walking (red on green). The misclassification occurs at the transition area between walking and nonwalking, which is reasonable since there are uncertainties for those points in the transition area.

Specifically, on Dataset \#7, where the training data and test data have no overlap but are all from the same 10 subjects, the proposed method achieves an accuracy of 90.22\%, which shows the effectiveness of the proposed method in separating walking data from nonwalking data. On Dataset \#8, where the training data and the test data are from different subjects, an accuracy of 85.57\% is obtained, which indicates that the proposed method has high generalization ability.

\subsection{Performance of LSTM at Different Data Settings}\label{sec:lstm-net-perf}
In this experiment, we examine how the network structure influences the performance of LSTM and how effective the different data settings are for classification.

\subsubsection{Different LSTM networks} The layers of LSTM can be constructed with information propagation in the forward direction only or in both the forward and backward directions. In this experiment, we test three LSTM network architectures, which are:
\begin{enumerate}[   $\vcenter{\hbox{\tiny$\bullet$}}$]
    \item SL-LSTM: an LSTM with one single hidden layer.
    \item Bi-LSTM: a bi-directional LSTM, with a layer forward and a layer backward.
    \item DL-LSTM: an LSTM with two hidden layers.
\end{enumerate}

\subsubsection{Different data settings}
The original data contain six channels. We investigate how the combination of channel(s) affects the performance. These data are constructed based on Dataset \#1, Dataset \#2, Dataset \#3 and Dataset \#4. Specifically, we build three network structures, with each evaluated on 8 different combinations of data channels, i.e., 4 for the interpolated data and 4 for the time-fixed data. The four interpolated data are:
\begin{enumerate}[ $\vcenter{\hbox{\tiny$\bullet$}}$]
  \item interp\_6: three-axis accelerometer data and three-axis gyroscope data of Dataset \#1 and Dataset \#2, sampled in an interpolation manner, with a data shape of 6$\times$128.
  \item interp\_acc: three-axis accelerometer data of Dataset \#1 and Dataset \#2, with a data shape of 3$\times$128.
  \item interp\_gyr: three-axis gyroscope data of Dataset \#1 and Dataset \#2, with a data shape of 3$\times$128.
  \item interp\_sqrt: the mean square root of the three-axis accelerometer data of Dataset \#1 and Dataset \#2, with a data shape of 1$\times$128.
\end{enumerate}
And the four time-fixed data are:
\begin{enumerate}[ $\vcenter{\hbox{\tiny$\bullet$}}$]
  \item fixed\_6: three-axis accelerometer data and three-axis gyroscope data of Dataset \#3 and Dataset \#4, sampled in a time-fixed manner, with a data shape of 6$\times$128.
  \item fixed\_acc: three-axis accelerometer data of Dataset \#3 and Dataset \#4, with a data shape of 3$\times$128.
  \item fixed\_gyr: three-axis gyroscope data of Dataset \#3 and Dataset \#4, with a data shape of 3$\times$128.
  \item fixed\_sqrt: the mean square root of the three-axis accelerometer data of Dataset \#3 and Dataset \#4, with a data shape of 1$\times$128.
\end{enumerate}

\begin{figure*}[!t]
    \begin{minipage}[t]{0.33\linewidth}
    \centering
    \includegraphics[height=5.1cm,width=5.1cm]{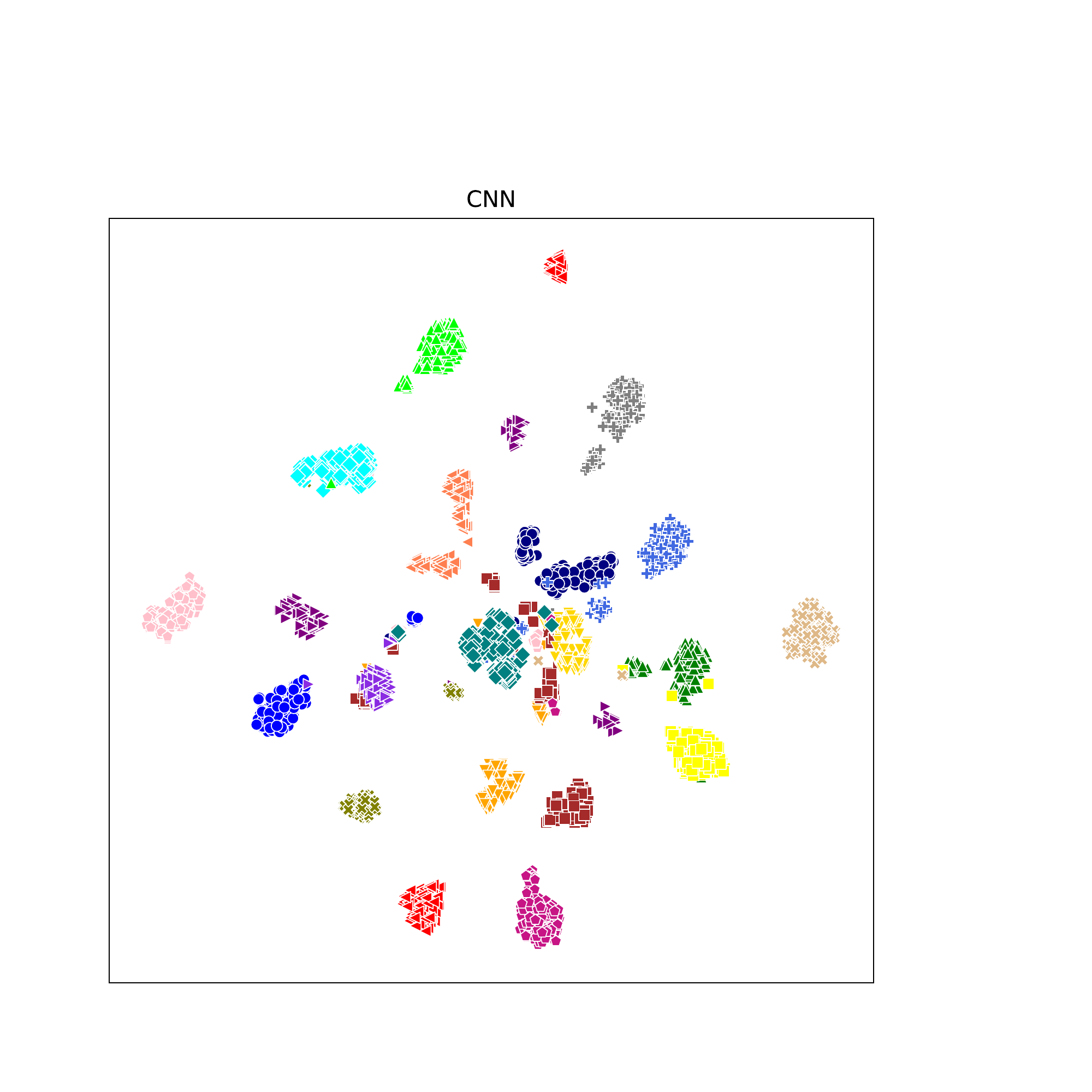}
    \centerline{(a)}
    \label{fig:data-1}
    \end{minipage}\hspace{0mm}
    \begin{minipage}[t]{0.33\linewidth}
    \centering
    \includegraphics[height=5.1cm,width=5.1cm]{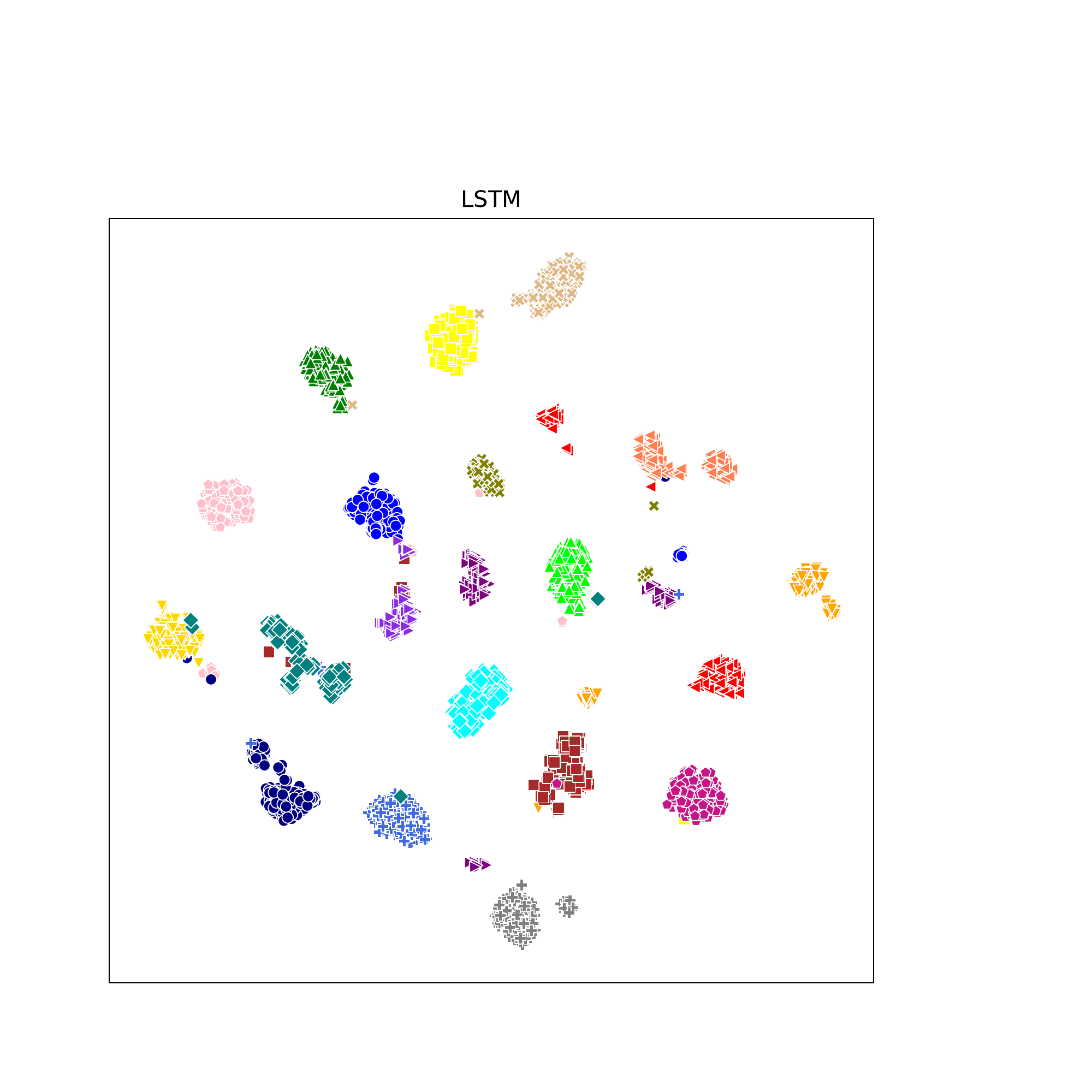}
    \centerline{(b)}
    \label{fig:data-2}
    \end{minipage}\vspace{0mm}
    \begin{minipage}[t]{0.33\linewidth}
    \centering
    \includegraphics[height=5.1cm,width=5.1cm]{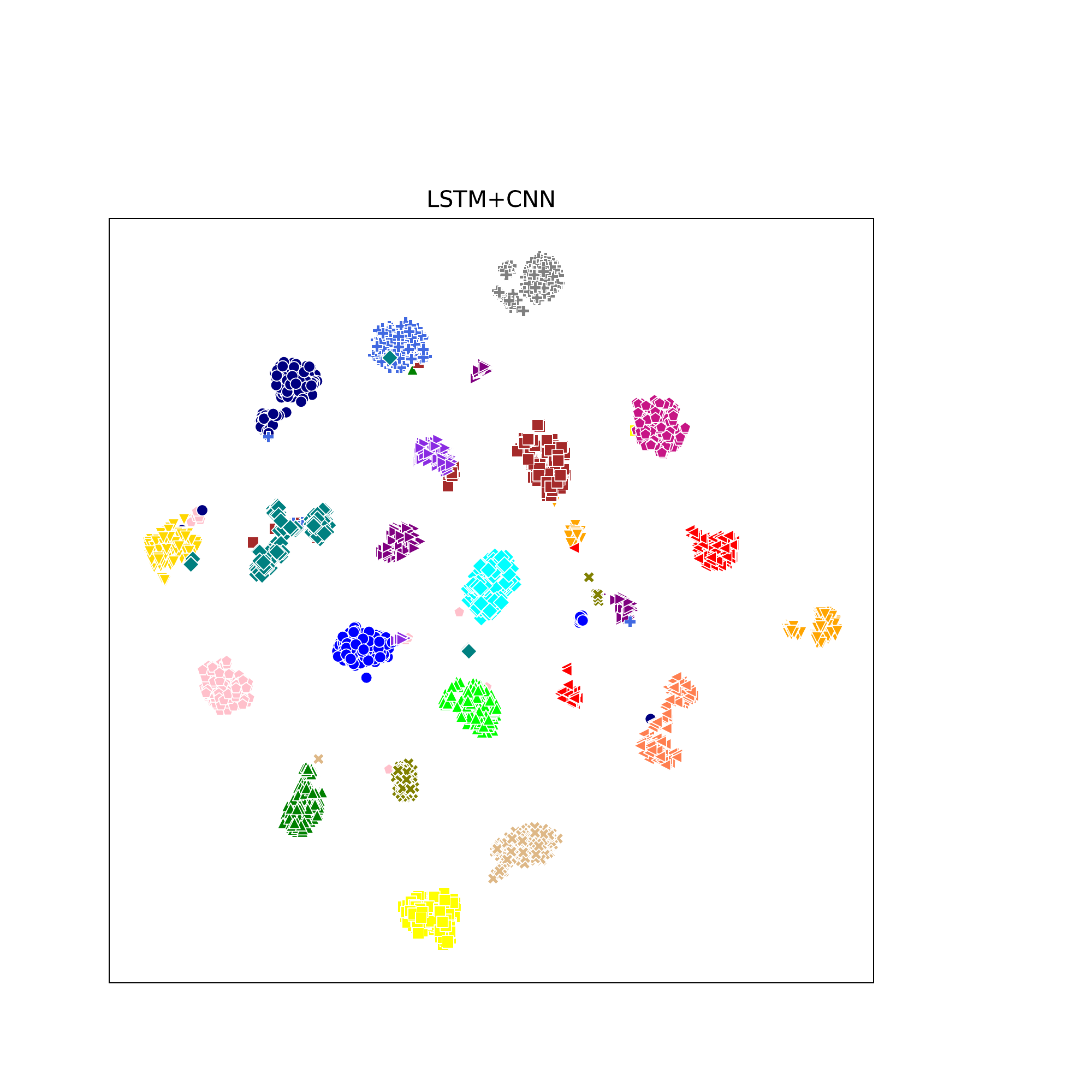}
    \centerline{(c)}
    \label{fig:data-3}
    \end{minipage}\vspace{-2mm}
    \caption{Visualizing the deep features of the test samples of 20 subjects in Dataset \#2 using t-SNE~\cite{maaten2008visualizing}. (a) LSTM. (b) CNN. (c)  `CNN+LSTM'. Features in visualization are from the full-collection layer of the three networks. Feature points for different subjects are marked in different colors and/or shapes. Note that, all the three models are trained on Dataset \#1. The t-SNE runs with a perplexity of 30, an embedding method of PCA, and a reduced dimension of 2.}
    \label{fig:t-SNE}
\end{figure*}

\subsubsection{Experimental details and results}
For the above LSTM networks,  the number of nodes in the hidden layer is set to 64. The last hidden layer is followed by a fully connected layer, which is used as a classification output layer. The size of the fully connected layer is 20 for the case of 20 subjects and 118 for the case of 118 subjects. All networks are trained with a learning rate of 0.0025 and 300 training epochs.

{Figure}~\ref{fig:lstms} shows the results of the three LSTM networks at eight different data settings. The data are constructed based on Dataset \#1 and Dataset \#3, which have 118 subjects. Figure~\ref{fig:lstms} shows that the results based on the accelerometer data are better than those based on the gyroscope data, and the results based on the gyroscope data are better than those based on the mean square root data. It indicates that, the accelerometer sensor is better than the gyroscope sensor at capturing the gait feature.

Figure~\ref{fig:lstms} also indicates that the results based on interpolated data are slightly better than those based on time-fixed data at the given settings. The best results on both the interpolated data and time-fixed data are obtained in the six-axis setting. Thus, the accelerometer data and the gyroscope data can complementary when representing gait features. Because DL-LSTM obtains the highest classification accuracy of 91.88\% at interp\_6, we use DL-LSTM and the interp\_6 data in our later experiments for comparison.

\subsection{User-Identification Performance}

In this subsection, we introduce several experiments to evaluate the performance of various methods for gait identification. As we consider a classification problem, we use accuracy as a metric to evaluate the performance. Accuracy is defined as,

{\small
\begin{equation}\label{eq:exp-metric}
\mathbf{Accuracy} = \frac{\mathbf{Correctly\ Classified\ Samples}}{\mathbf{Total\ Testing\ Samples}}.
\end{equation}}

\subsubsection{Comparison methods} To evaluate the proposed methods for person identification, ten methods are included for comparison:
\begin{enumerate}[ $\vcenter{\hbox{\tiny$\bullet$}}$]
    \item Fourier~\cite{mantyjarvi2005identifying}: first, the gait data are processed by an autocorrelation operation; then, the results are converted into the frequency domain via FFT; finally, the first 40 FFT coefficients per channel are selected as the gait features, and a one-vs-all SVM is employed for classification.

    \item Wavelet~\cite{XuCMU12btas}: the gait data are decomposed using the Mexican hat wavelet, and the low-frequency component of the results is used as the gait feature for person identification. A one-vs-all SVM classifier is employed.

    \item EigenGait~\cite{zou2017robust}: the inertial gait data are decomposed in the eigen space, and the principle components are taken as gait features for person identification. A one-vs-all SVM classifier is employed.

    \item LSTM: the DL-LSTM introduced in Section~\ref{sec:lstm-net-perf} that is trained from scratch.

    \item CNN: the CNN introduced in Section~\ref{sec:iden-net} that is trained from scratch.

     \item CNN+LSTM: the network introduced in Figure~\ref{fig:net-iden}, which combines the above two networks. The whole network is trained from scratch.

     \item CNN$_{fix}$+LSTM: the network introduced in Figure~\ref{fig:net-iden}. The CNN is fixed with pre-trained weight parameters, and the LSTM and fully connected layer are trained from scratch.

     \item CNN+LSTM$_{fix}$: the network introduced in Figure~\ref{fig:net-iden}. The LSTM is fixed with pre-trained weight parameters, and the CNN and fully connected layer are trained from scratch.

     \item IdNet~\cite{gadaleta2018idnet}: a CNN-based person-identification method using inertial data collected by a smartphone. The IdNet method is trained from scratch. The dataset, named IdNet dataset, is constructed from 50 subjects, with 15,096 samples for training and 6,471 samples for test.

     \item DeepConvLSTM~\cite{ordonez2016deep}: a deep-learning framework composed of convolutional and LSTM recurrent layers. It is capable of automatically learning feature representations and modeling the temporal dependencies between their activation. The model is trained from scratch.
\end{enumerate}

\begin{table}[!t]
  \centering   \small
  \caption{Classification performance of traditional methods}\vspace{-2mm}
    \begin{tabular}{|c|c|c|c|c|}
    \hlinew{1.5pt}
    \multicolumn{1}{|c|}{\textbf{Dataset}} & \multicolumn{1}{|c|}{\textbf{Channels}} & \multicolumn{1}{c|}{\textbf{EigenGait}} & \multicolumn{1}{c|}{\textbf{Wavelet}} & \multicolumn{1}{c|}{\textbf{Fourier}} \\
    \hlinew{1.2pt}
    \hline
    \hline
    \multirow{3}[6]{*}{\tabincell{c}{Dataset \#2\\ (20 subj.) \\ \ }} & interp\_6 & 87.03\% & 87.20\% & 93.64\% \\
          \cline{2-5}
          & interp\_acc & 86.95\% & 88.84\% & 90.15\% \\
          \cline{2-5}
          & interp\_sqrt & 65.24\% & 78.40\% & 66.90\% \\
    \hline \hline
    \multirow{3}[6]{*}{\tabincell{c}{Dataset \#1\\ (118 subj.) \\ \ }} & interp\_6 & 76.58\% & 75.13\% & 81.55\% \\
    \cline{2-5}
          & interp\_acc & 76.66\% & 78.96\% & 75.24\% \\
          \cline{2-5}
          & interp\_sqrt & 37.62\% & 52.25\% & 35.72\% \\
\hlinew{1.5pt}
    \end{tabular}%
  \label{tab:all-compare-svm}%
\end{table}%

\begin{table*}[!t]
  \centering   \small
  \caption{Classification performance of deep learning-based methods}\vspace{-2mm}
    \begin{tabular}{|c|c|c|c|c|c|}
    \hlinew{1.5pt}
    \textbf{Classification Methods} & \tabincell{c}{\textbf{Dataset \#1}\\ \textbf{(118 Subjects)}} & \tabincell{c}{\textbf{Dataset \#2}\\ \textbf{(20 Subjects)}}& \tabincell{c}{\textbf{IdNet Dataset}\\ \textbf{(50 Subjects)}} & \tabincell{c}{\textbf{{OU-ISIR Dataset}}\\ \textbf{(745 Subjects)}} & \tabincell{c}{\textbf{{OU-ISIR Dataset}}\\ \textbf{(745, finetune)}} \\
    \hlinew{1.2pt}
    \hline
    \hline
    IdNet~\cite{gadaleta2018idnet} & 92.91\% & 96.78\% & 99.58\% & 44.29\% &46.20\%\\
    \hline
    DeepConvLSTM~\cite{ordonez2016deep} &  92.25\% & 96.80\% &  99.24\% &  37.33\% &  38.32\%   \\ 
    \hline
    LSTM   & 91.88\% & 96.98\% & 99.46\%  & \textbf{66.36}\% &  \textbf{72.32}\%  \\
    \hline
    CNN  & 92.89\% & 97.02\% & 99.71\% & 40.60\%& 45.14\%  \\
    \hline
    CNN+LSTM & 92.51\% & 96.82\% & 99.61\%  & 34.28\%& 35.96\% \\
    \hline
    CNN$_{fix}$+LSTM & 92.94\% & 97.04\% & 99.64\% & - & -\\
    \hline
    CNN+LSTM$_{fix}$ & \textbf{93.52}\% & \textbf{97.33}\% & \textbf{99.75}\%  & - & -\\
    \hlinew{1.5pt}
    \end{tabular}%
  \label{tab:all-classify}%
\end{table*}%

The Fourier, Wavelet and EigenGait methods were chosen for comparison because they are all commonly used for time-series signal analysis and feature extraction in the gait recognition. The IdNet and DeepConvLSTM were chosen for an extensive comparison because they are two representative deep learning-based methods.

\subsubsection{Performance of traditional methods}
The classification results of the three traditional methods, i.e., Fourier, Wavelet and EigenGait, are shown in Table~\ref{tab:all-compare-svm}. The Fourier-transform-based method performs the best on both Dataset \#1 and Dataset \#2, where the accuracy is 81.55\% and 93.64\%, respectively. These best performance is obtained on the interpolation data with six-axis inertial inputs.

\subsubsection{Performance of deep learning methods}
Three datasets, i.e., Dataset \#1, Dataset \#2 and the IdNet dataset, are used to evaluate the seven deep learning-based methods: LSTM, CNN, CNN+LSTM, CNN$_{fix}$+LSTM, CNN+LSTM$_{fix}$, IdNet and DeepConvLSTM. The classification results are shown in Table~\ref{tab:all-classify}. All seven deep learning-based methods achieve greater than 91.8\% accuracy on Dataset \#1 (with 118 subjects) and greater than 96.7\% accuracy on Dataset \#2 (with 20 subjects), much higher than that obtained by the traditional methods. On the IdNet dataset, all the seven methods achieve greater than 99.2\% accuracy, and very little difference is observed in the classification results. This is because the data in the IdNet dataset are collected from relatively standard walking style, which makes classification less challenging.
To illustrate the effectiveness and generalization of feature extraction, we visualize the extracted features in Figure~\ref{fig:t-SNE}. Here, the CNN and LSTM models were trained on Dataset \#1, and the test samples for feature visualization were from Dataset \#2. It can be observed that, the feature points in Figure~\ref{fig:t-SNE}(b) are closer for the same class and farther away from each other for different classes than those in Figure~\ref{fig:t-SNE}(a). This result indicates that the CNN features are more discriminative than the LSTM features. We can also see that the distributions of the feature points in Figure~\ref{fig:t-SNE}(b) and Figure~\ref{fig:t-SNE}(c) are quite similar, and this is consistent with the results on Dataset \#1 in Table~\ref{tab:all-classify}, indicating that the CNN features hold similar discriminative power to those of the hybrid network.

Table~\ref{tab:all-classify} also shows that CNN performs better than LSTM on both Dataset \#1 and Dataset \#2. For the more challenging 118-subject case, CNN outperforms LSTM by approximately 1\% in terms of accuracy; thus, the CNN can better extract the gait features from the inertial gait curves.  The results show that the CNN+LSTM trained from scratch is not guaranteed to outperform the CNN or LSTM. One possible reason is that the CNN and LSTM are in parallel in the identification network, and the hybrid network is more complex and has not been fully trained. That is to say, during the training process, the hybrid network trained from scratch may suffer from overfitting.

When fixing the weight parameters of one network and training the other, we obtain improved performance over that of single-network-based methods on Dataset \#1 and Dataset \#2. This is because the final loss directly reflects the status of the unfixed network, and the two types of features are complementary in the classification. Moreover, CNN+LSTM$_{fix}$ achieves the best performance and outperforms CNN$_{fix}$+LSTM by approximately 0.6\% and 0.3\% on Dataset \#1 and Dataset \#2, respectively. This may be because it is much more difficult to train the LSTM than the CNN in a parallel-structured network. IdNet obtains an accuracy approximately 0.6\% and 0.5\% lower than those of CNN+LSTM$_{fix}$ on Dataset\#1 and Dataset\#2, respectively. While for DeepConvLSTM, the accuracy is approximately 1.2\% and 0.5\% lower.

We also evaluate the deep learning-based methods on the OU-ISIR dataset~\cite{iwama2012isir,Ngo14pr}, which was collected from 745 subjects. However, there are few inertial gait data for any single subject (18.73 seconds at most and 5.61 seconds at least with a sensor frequency of 100Hz). We include this dataset to examine the performance of deep neural networks in the scenarios with limited training samples and large numbers of subjects. We sample the data from the inertial sequence with a length of 128 time points and an interval of 50. Thus, the dimension of a data sample is 6$\times$128. Considering that the minimum number of data samples for any single subject is 8, we partition the samples for training and testing at a ratio of 7:1. Thus, we obtain 13,212 samples for training and 1,409 samples for test. We then train the deep learning-based methods by two means: one model is trained from scratch, and the other is pretrained on Dataset \#1 and then fine-tuned on the OU-ISIR dataset. The results are shown in Table~\ref{tab:all-classify}. As can be seen, the LSTM obtains the best performance on the OU-ISIR dataset. One possible reason may be that the LSTM has the lightest-weight architecture among the networks. Consequently, the limited number of samples can support the training of LSTM relatively better than trainings of other more complex networks. It is also observed that all the methods achieve improved performance after fine-tuning, but their overall performance is below 73\%.

\subsection{User-Authentication Performance}

\subsubsection{Comparison methods} Eleven methods are included for comparison in the authentication experiments:
\begin{enumerate}[ $\vcenter{\hbox{\tiny$\bullet$}}$]
     \item Fourier~\cite{mantyjarvi2005identifying}: the first 80 FFT coefficients per channel are selected in Dataset \#5 as the gait features, while in Dataset \#6, 40 per channel are selected.  A 2-class SVM is employed as a classifier.

    \item Wavelet~\cite{XuCMU12btas}: the continuous wavelet transform is employed to obtain the components from scale 1 to 20. The energy of the frequency band signal at scale $i$ is ${E}_{i}=({\sum_{k=1}^M\vert{x_i(k)}\vert^{2}})^{\frac{1}{2}}$,
        where $x_i(k)$ denotes the discrete point amplitude of the reconstructed signal at scale $i$ and $M$ denotes the number of discrete points. In our case, $M$=256 is for the horizontally aligned, and $M$=128 is for the vertically aligned. We construct the feature vector as
         \begin{equation}\label{eq:feature_vectors}
        \begin{aligned}
        \textbf{F}&=[\frac{E_1}{E},\frac{E_2}{E},...,\frac{E_{20}}{E}],\\
        \end{aligned}
        \end{equation}
        with ${E}$=$({\sum_{i=1}^{20}{E_i}^{2}})^{\frac{1}{2}}$, which places an L2 normalization. A 2-class SVM classifier is employed.

    \item EigenGait~\cite{zou2017robust}: the principle components generated by the eigen decomposition are taken as gait features for person authentication. A 2-class SVM classifier is employed.
    \item CNN\_horizontal: the CNN network using horizontally aligned data pairs as the input.
    \item CNN\_vertical: the CNN network using vertically aligned data pairs as the input.
    \item LSTM\_horizontal: the LSTM network using horizontally aligned data pairs as the input.
    \item LSTM\_vertical: the LSTM network using vertically aligned data pairs as the input.
    \item CNN+LSTM\_horizontal: the `CNN+LSTM' network, as have been introduced in Figure~\ref{fig:net-auth}, using horizontally aligned data pairs as the input.
    \item CNN+LSTM\_vertical: the `CNN+LSTM' network using vertically aligned data pairs as the input.
    \item CNN$_{fix}$+LSTM\_horizontal: the `CNN$_{fix}$+LSTM' network, as have been introduced in Figure~\ref{fig:net-auth}, using horizontally aligned data pairs as the input. The weight parameters of CNN are fixed.
    \item CNN$_{fix}$+LSTM\_vertical: the `CNN$_{fix}$+LSTM' network using vertically aligned data pairs as the input. The weight parameters of CNN are fixed.
\end{enumerate}

Note that, CNN$_{fix}$ is pretrained with data from 98 subjects in Dataset \#1. These 98 subjects are the same subjects used for training in Dataset \#5 and Dataset \#6. As a result, the subjects of the test samples in Dataset \#5 and Dataset \#6 are unseen to the CNN$_{fix}$, which makes the authentication task challenging.

\subsubsection{Metric}
Accuracy is employed as a metric to evaluate the performance of various methods. Moreover, the ROC curve is employed for comparison. The ROC curve is obtained by plotting the true positive rate (TPR) against the false positive rate (FPR) at varying threshold settings. The TPR and FPR are defined as,

{\small
\begin{equation}\label{eq:exp-tpr}
\mathbf{TPR} = \frac{\mathbf{True\ Positive}}{\mathbf{True\ Positive +\ False\ Negative}},
\end{equation}}\vspace{-2mm}
{\small
\begin{equation}\label{eq:exp-fpr}
\mathbf{FPR} = \frac{\mathbf{False\ Positive}}{\mathbf{False\ Positive +\ True\ Negative}}.
\end{equation}}

\subsubsection{Performance} For all authentication methods except EigenGait{\footnote{EigenGait can only decompose feature vectors with a predefined dimension.}}, the input data can be aligned in two manners:  horizontally or vertically. For example, with two 6$\times$128 samples, the input data can be 6$\times$256 when aligned horizontally or 12$\times$128 when aligned vertically. Dataset \#5 and Dataset \#6 are constructed from the same 118 subjects and the same samples. The only difference is, samples in Dataset \#5 are horizontally aligned and in Dataset \#6 are vertically aligned.

\begin{table}[!t]
  \small \center
  \caption{Authentication Performance (in Accuracy)}\vspace{-2mm}
    \begin{tabular}{|c|c|c|}
    \hlinew{1.5pt}
    \textbf{Authentication Methods} & \tabincell{c}{\textbf{Dataset \#5}\\ \textbf{(Horizontal)}} & \tabincell{c}{\textbf{Dataset \#6}\\ \textbf{(Vertical)}}\\
    \hlinew{1.2pt}
    \hline
    \hline
    CNN & 78.47\% & 87.72\% \\
    \hline
    LSTM & 82.39\% & 91.70\% \\
    \hline
    CNN+LSTM & 84.45\% &92.79\% \\ 
    \hline
    CNN$_{fix}$+LSTM & 85.54\% & \textbf{93.75}\% \\
    \hline \hline
    EigenGait& - & 78.97\%\\
    \hline
    Wavelet&  78.55\%   & 77.37\%  \\
    \hline
    Fourier& \textbf{92.70}\% & 61.86\%   \\
    \hlinew{1.5pt}
    \end{tabular}\label{tab:authentication6}
\end{table}%

First, we must determine which alignment strategy results in better performance. Table~\ref{tab:authentication6} shows the authentication results obtained by four deep learning-based methods, i.e., LSTM, CNN, CNN+LSTM and CNN$_{fix}$+LSTM, and three traditional methods, i.e., EigenGait, Wavelet and Fourier.

\begin{table}[!t]
  \centering   \small
  \caption{Authentication performance on OU-ISIR dataset}\vspace{-2mm}
    \begin{tabular}{|c|c|c|}
    \hlinew{1.5pt}
    \textbf{Authentication Methods}  & \tabincell{c}{\textbf{Horizontal}} & \tabincell{c}{\textbf{Vertical}} \\
    \hlinew{1.2pt}
    \hline
    \hline
    CNN & 97.87\% & 98.33\%  \\
    \hline
    LSTM & 96.07\% & 96.60\%  \\
    \hline
    CNN+LSTM & \textbf{99.33}\% & \textbf{99.60}\%  \\
    \hlinew{1.5pt}
    \end{tabular}%
  \label{tab:auth-OU-ISIR}%
\end{table}%

For the deep learning-based methods, the results obtained from vertically aligned data are much better than those from horizontally aligned data. The improvements are approximately 9\%, 9\%, 8\% and 8\% for the CNN, LSTM, CNN+LSTM and CNN$_{fix}$+LSTM, respectively. One possible reason for the difference is that a pair of samples aligned vertically are aligned along the time dimension. Since the input data are time series, aligning data along the time dimension facilitates the comparison for both the LSTM and the 1-D CNN. Moreover, LSTM achieves better results than CNN, which indicates that the LSTM can better handle time-series data by associating the past and upcoming signals for feature learning and state prediction. Additionally, the CNN and LSTM are found to be complementary in further improving the performance. In Table~\ref{tab:authentication6}, CNN+LSTM and CNN$_{fix}$+LSTM substantially outperform the stand-alone CNN or LSTM, which demonstrates the effectiveness of the proposed fusion strategy. Moreover, CNN$_{fix}$+LSTM shows superior performance to that of CNN+LSTM, possibly because the CNN block is harder to train in the relatively more complex CNN+LSTM network than in the stand-alone CNN network. The dataset collected on 118 subjects produces good results but is still not sufficient for training a satisfactory and general model. Thus, CNN+LSTM trained from scratch is more likely to overfit than CNN$_{fix}$+LSTM.

For the three traditional methods, all of them produce much worse results than the deep learning-based methods on the vertically aligned data. However, on the horizontally aligned data, the Fourier-transform-based method achieves the highest accuracy 92.70\% among all methods. This accuracy is much higher than 61.86\%, which is obtained on vertically aligned data. This result demonstrates the potential of frequency-domain transformation to effectively capture the discriminative characteristics of concatenated gait time series. The feature vector constructed via Fourier transform  samples the coefficients of the frequencies from low to high in the frequency domain. For horizontally aligned data, feature vectors can be constructed based on the fused signal, whereas for vertically aligned data, feature vectors can only be constructed independent from each other. The independent features are less capable of capturing discriminative characteristics for authentication. For the wavelet-based method, the results on the horizontally and vertically aligned data are similar, which indicates that the channel-combination method for a pair of samples has little effect on the components of the continuous wavelet transform at different scales.

\begin{figure}[!t]
  \centering \vspace{1mm} \hspace{-5mm}
  \includegraphics[width=0.8\linewidth]{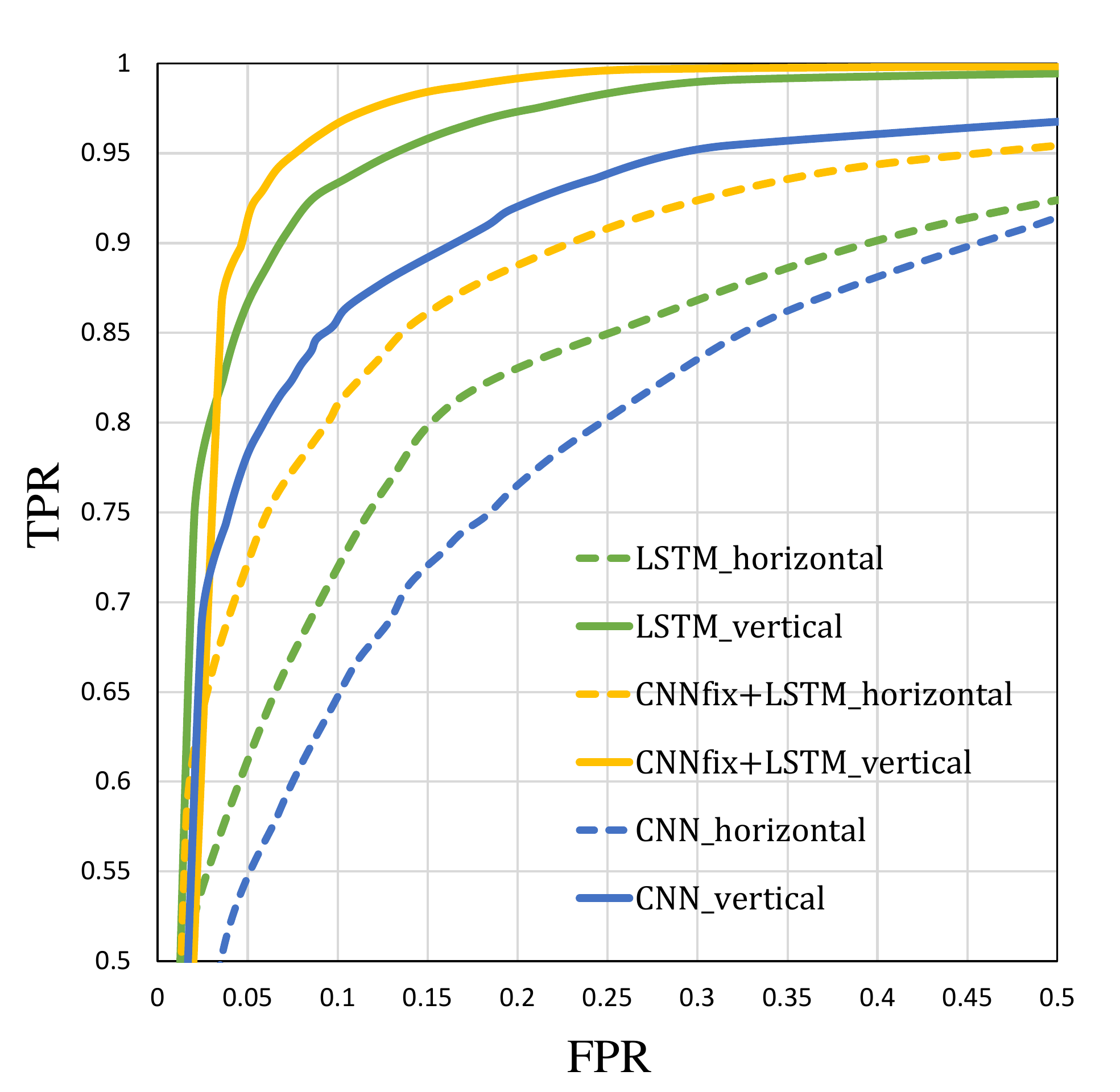} \vspace{-2mm}
  \caption{ROC curves of six deep learning-based authentication methods.}
  \label{fig:roc}
\end{figure}

Figure~\ref{fig:roc} shows the ROC curves obtained by three deep-learning authentication methods using two different data-alignment strategies. The three deep learning methods are plotted in three different colors, and the different data are indicated by solid lines and dash lines. The deep learning methods perform much better on vertically aligned data than on horizontally aligned data, which indicates that the vertically aligned inertial data can better fit the neural networks for authentication. In the vertically aligned data, the gait phases of two input samples that are close in time will also be physically close. With regarding to the signal-processing manner of CNN and LSTM, this property can facilitate the discrimination of the difference between two samples. Also, we can see that the proposed hybrid network obtains higher performance than the stand-alone ones.

We further evaluate the proposed method on the OU-ISIR dataset~\cite{Ngo14pr}. For authentication, we first resample the inertial data from 100Hz to 50Hz and then divide the data into two parts: one with 670 subjects, and the other with 75 subjects. Specifically, we construct 13,400 data pairs from the 670 training subjects and 1,500 data pairs from the 75 testing subjects. Each data pair has two versions, i.e., horizontally aligned (6$\times$256) and vertically aligned (12$\times$128). The number of positive and negative samples is equal in both the training data and the test data. As can be seen from Table~\ref{tab:auth-OU-ISIR}, all three deep learning-based methods achieve authentication accuracies above 96\%. CNN+LSTM achieves an improved performance over the stand-alone CNN and LSTM models, further indicating the effectiveness of the proposed hybrid neural network.

\section{Discussion} \label{sec:discussion}
Indeed, some exiting methods already integrated CNN and RNN for video processing and achieved high performance in various tasks, e.g., frame prediction, action recognition, etc. However, these methods handle a sequence of images rather than inertial signals. For the inertia-based gait recognition, we proposed a new and reasonable way for seamlessly integrating the CNN and RNN. In our network, the CNN is used to extract two-dimensional features of gait time series. One of these dimensions is the temporal feature, while the other dimension here is the spatial feature, as regarding to the fact that the corresponding signals of different sensors are collected simultaneously. The spatial property of the data captured by different sensors is rarely considered in the traditional sense. We carefully designed a CNN with a set of one-dimensional convolutional kernels to process the spatial features. Moreover, in the authentication network, the feature data processed by the CNN still retain the temporal dimension, which can be further processed by the LSTM as a temporal feature.

For the reasonability of constructing two-dimensional time series, we motivate it under the assumption that the signals of accelerometer data in axis X, Y and Z at the same time stamp are relevant, and the signals of accelerometer and gyroscope are also relevant. In order to embed these related information into the deep model, we arrange the tri-axis accelerometer and gyroscope gait data into six-axis inertial gait data. One-dimensional convolution operations can be applied to process the information among different axes. As shown in Figure~\ref{fig:net-auth} and Table II, we use a sequence of one-dimensional kernels to do the convolution, that is, 1$\times$9, 1$\times$3 and 1$\times$3 for the first three convolutions, where the signals are separately processed along the time zone. As the convolution is a local operation, the output features are supposed to also hold the property of time series. Then, a 6$\times$1 kernel is applied to the previous convolution results and spatially correlates the signals on different axes, and produce 1$\times$16$\times$128 data. Note that, 16 is the number of the features along the time axis. After re-arranging it into 128$\times$16, we also get the time series. The proposed one-dimensional convolution kernels can guarantee the temporal property of the final convolution results. Consequently, we can feed the convolution output into an LSTM for further inference.

We do not use the authentication network for identification. This is because the authentication network does not achieve high performance in the identification task in our experiments. In the authentication, the sequentially combined network `CNN+LSTM' is more complex than a single CNN or RNN and thus more difficult to train, which commonly requires a larger amount of training data to avoid overfitting. However, in our datasets, the training samples for identification is much less than that for authentication, e.g., about 20,000 v.s. 60,000. That is a possible reason why the `CNN+LSTM' in authentication cannot achieve high performance in the identification task. In the future work, we will study to simplify the sequential `CNN+LSTM' to handle the identification task.

It is worth noting that the experimental results are obtained on Datasets \#1-\#8 collected from 118 healthy persons. Therefore, the datasets have limitations for the use of performance evaluation on people with physical problems. Also note that the triplet and contrast losses both have been proven to be very effective for training neural models of gait recognition~\cite{takemura2017csvt}. In such cases, the number of training triplets is quadratically increased w.r.t. the number of original training samples. This approach may relieve the problem of training data shortage, which we leave as our future work.

In large-scale person identification, data collection may suffer from privacy issues. In this case, the data collector and the smartphone users should reach a prior agreement on data usage (e.g., smartphone users should sign an authorization form before logging into a data collector application on the smartphone). Note that we do not constrain the users to change smartphones. We only assume that a given smartphone is held by one person at a time. In other words, everyone is allowed to hold multiple smartphones, and any smartphone may be used by different persons. Person identification, however, will not be affected even if users change smartphones.

\section{Conclusion} \label{sec:conclusion}
In this paper, gait recognition using smartphones in the wild was studied. A hybrid method was proposed to seamlessly combine the DCNN and DRNN for robust inertial gait feature representation. During gait data collection, the smartphones were used under unconstrained conditions, and information about when, where, and how the user walks was totally unknown. A fully convolutional neural network was presented to partition the inertial data into walking and nonwalking sessions, and hierarchical convolutional features were fused for accurate semantic segmentation. Then, a CNN was proposed to transform the input time series into convolutional feature maps. With specially designed one-dimensional convolutional kernels, the convolutional output still retained the property of time series. An LSTM was employed to further process the time-series features for gait recognition. Two major datasets were collected for performance evaluation.

Based on the experimental results, we can draw the following conclusions: i) the performance based on accelerometer data is generally better than that based on gyroscope data, and the accelerometer data and gyroscope data can be complementary, further improving the performance; ii) the CNN and LSTM are both very effective for gait feature extraction from inertial time series; iii) in person identification, the data separated in steps provide slightly better performance for LSTMs than that separated in a fixed time span; iv) in person authentication, the inertial data aligned vertically provide much better results than that aligned horizontally for the DCNN and LSTM; and v) the proposed hybrid network obtained significantly improved performance over the stand-alone ones. To promote the research of inertia-based gait recognition, we have shared the datasets, codes and the trained model at {\color{blue} \url{https://github.com/qinnzou/}.}

\bibliographystyle{IEEEtran}
\bibliography{refs-arxiv}
\vspace{2mm}


\end{document}